\documentclass[journal]{IEEEtran}

\usepackage[colorlinks,linkcolor=blue,anchorcolor=blue,citecolor=green,bookmarks=true]{hyperref}
\usepackage{graphicx}
\usepackage{dsfont}
\usepackage{stfloats}
\usepackage{subfigure}
\usepackage{amsmath}
\usepackage{cite}
\usepackage[table,xcdraw]{xcolor}
\usepackage{booktabs}
\usepackage{algorithm}
\usepackage{algorithmic}
\usepackage{amssymb}
\usepackage{multirow}
\usepackage{listings}
\usepackage{tikz}
\definecolor{colorchange}{rgb}{0.1,0.8,0.1}
\begin{document}
\title{Multi-Temporal Scene Classification and Scene Change Detection with Correlation based Fusion}

\author{Lixiang~Ru,
	Bo~Du,~\IEEEmembership{Senior Member, IEEE},
	Chen~Wu,~\IEEEmembership{Member, IEEE}
	\thanks{This work was supported in part by the National Natural Science Foundation of China under Grant 61971317, 41801285, 61822113 and 41801285. \textit{Corresponding author: Bo Du and Chen Wu.}}
	\thanks{L. Ru and B. Du are with the School of Computer Science, Wuhan University, Wuhan 430072, China. (e-mail: rulixiang@whu.edu.cn, gunspace@163.com)}
	\thanks{C. Wu is with the State Key Laboratory of Information Engineering in Surveying, Mapping and Remote Sensing, Wuhan University, Wuhan 430072, China. (e-mail: chen.wu@whu.edu.cn)}
}

\markboth{XXXX}%
{Ru \MakeLowercase{\textit{et al.}}: Multi-Temporal Scene Classification and Scene Change Detection with Correlation based Fusion}
\maketitle

\begin{abstract}
	Classifying multi-temporal scene land-use categories and detecting their semantic scene-level changes for imagery covering urban regions could straightly reflect the land-use transitions. Existing methods for scene change detection rarely focus on the temporal correlation of bi-temporal features, and are mainly evaluated on small scale scene change detection datasets. In this work, we proposed a CorrFusion module that fuses the highly correlated components in bi-temporal feature embeddings. We firstly extracts the deep representations of the bi-temporal inputs with deep convolutional networks. Then the extracted features will be projected into a lower dimension space to computed the instance-level correlation. The cross-temporal fusion will be performed based on the computed correlation in CorrFusion module. The final scene classification are obtained with softmax activation layers. In the objective function, we introduced a new formulation for calculating the temporal correlation. The detailed derivation of backpropagation gradients for the proposed module is also given in this paper. Besides, we presented a much larger scale scene change detection dataset and conducted experiments on this dataset. The experimental results demonstrated that our proposed CorrFusion module could remarkably improve the multi-temporal scene classification and scene change detection results.
\end{abstract}

\IEEEpeerreviewmaketitle

\section{Introduction}
\IEEEPARstart{W}{ith} the continuous evolution of remote sensing technologies, more earth observations imagery with higher spatial resolution by airborne or spaceborne sensors is now produced every day. Compared with low and medium spatial resolution imagery, high spatial resolution imagery exhibits much more detailed contextual and texture information of landscapes, which makes it possible to perform scene-level land use and land cover analysis, such as scene classification \cite{cheng2017remote,Wang2020}, scene segmentation \cite{Hazel2000,Luo2020} and targeted object detection \cite{ElMikaty2017,Tao2019}.
\par Among them, remote sensing scene classification, which aims to assign semantic labels to a query of remote sensing scene images, has been a very hot topic in recent years. There have been numerous works on scene classification using various methods. In \cite{Sridharan2015}, \textit{Sridharan et al.} used a bag of words model \cite{wallach2006topic} and line features for scene classification. In \cite{Zhang2016}, \textit{Zhang et al.} ensembled multiple DNN models with gradient boosting to categorize image scene effectively. \cite{Zheng2019} introduced multi-scale pooling and Fisher Vector method to enhance the discriminability of learned features. A recent work proposed a densely connected CNN model with attention based multiple instance pooling for scene classification \cite{Bi2020}.
\par
\begin{figure}[]
	\centering
	\includegraphics[scale=0.45]{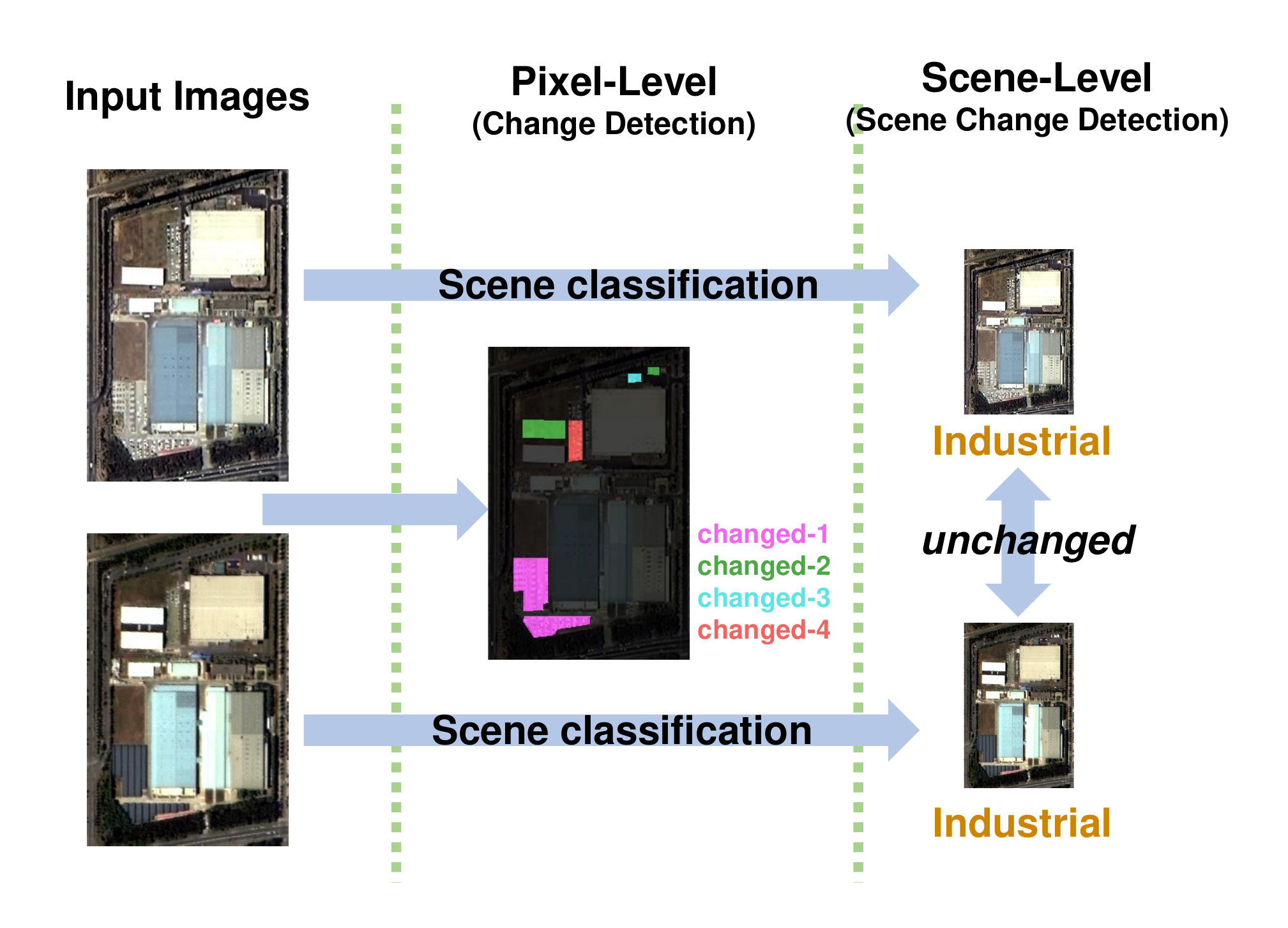}
	\caption{The difference between pixel-level change detection and scene change detection.}
	\label{fig_scd}
\end{figure}

 However, these existing scene classification works mainly concern mono-temporal images and rarely pays attention to classifying multi-temporal images. For multi-temporal scene images, there have been numerous works about detecting the pixel-level and object-level changes or further identifying their change types \cite{Saha2019, tang2013fault,khan2017forest}. However, as shown in Fig~\ref{fig_scd}, the detected pixel- or object-level changes cannot reflect the changes at land use and land cover scene-level, such as the Bare Land to Residential Region change. Since land-use scene gives an intuitive interpretation of given urban regions, detecting changes at scene-level will directly provide the transition information and helps to further urban planning.
\par In literature \cite{Wu2016}, scene change detection is defined as classifying multi-temporal scene images and comparing their changes at semantic scene-level, and it's drawing attention in recent years \cite{ru2019,du2018}. The work in \cite{Wu2016} provided a scene change detection framework for multi-temporal high-resolution imagery based on Bag of Visual Words (BoVW) model \cite{yang2007evaluating}. This framework utilizes BoVW model to encode multi-temporal scene images, and employs SVM classifier to obtain the scene classification results. The scene change detection results are obtained by post-classification comparison. However, this framework didn't take much consideration of the temporal correlation between the images acquired in the same location but different time. In a further work, this framework is improved using Kernel Slow Feature Analysis (KSFA) and Bayesian Fusion \cite{Wu2017}. This new method still takes BoVW to perform feature encoding and representation, and uses kernel SVM classifier to compute the classification probabilities as the initial results. Then the change probability of the bi-temporal images is computed with KSFA. Finally, the Bayesian Fusion is utilized to maximize the posterior probabilities based on all these computed probabilities. The results demonstrated that temporal correlation could remarkably boost the classification and change detection results. Still, methods in \cite{Wu2016,Wu2017} are both based on shallow handcrafted features, which're considered to be not effective in feature representation for large-scale dataset. Moreover, their handcrafted designs also determine the different modules of them could cannot be jointly optimized in an end-to-end way.
\par In the past several years, deep convolutional neural network (CNN) had been proposed and applied to diverse domains, including image classification, semantic segmentation, object detection and etc. \cite{Krizhevsky2012, Long2015,Girshick_2014_CVPR}. It had been showed that CNN could work brilliantly in remote sensing imagery related tasks such as scene classification and recognition \cite{nogueira2017towards, he2019skip, ghosh2018stacked}. Therefore, it's a natural choice to perform multi-temporal scene change detection with CNN. In literature \cite{Wang2019}, \textit{Wang et al.} proposed an end-to-end scene change detection network. This work firstly takes the bi-temporal scene images as inputs to extract convolutional feature representations, then the softmax classifier is employed for categorization. A Deep Canonical Correlation Analysis (DCCA) \cite{Andrew2013} regularization term is utilized in the objective function to maximize the correlation between the unchanged scene image pairs.
\par However, Since all previous scene change detection tasks are performed on small scale datasets, DCCANet doesn't show much superiority to BoVW based methods. Both of them are also very likely to overfit the training set and thus achieve poor generalization performance on testing set. Moreover, in the optimization process of DCCANet, DCCA term is optimized using minibatch gradient descent algorithms. However, it had been demonstrated in \cite{Wang2015,Wang2016b}, that DCCA could not be reliably optimized by a minibatch based optimization algorithms in the original formulation in \cite{Andrew2013}. To solve this problem, in \cite{Chang2018a}, \textit{Chang et al.} proposed Soft DCCA as an efficient equivalent of DCCA. Nevertheless, Soft DCCA still focuses on learning correlated features of multi-view inputs and doesn't utilize the correlated features to enhance the feature representation abilities.
\par In this paper, we proposed a correlation based feature fusion module called CorrFusion for multi-temporal scene classification and scene change detection. In this work, we starts with extracting the deep latent representations of bi-temporal input scene images using two independent convolutional modules. Then the extracted features are respectively projected into a new feature space with fully connected layers. We designed a feature fusion module based on the temporal correlation between the multi-temporal images calculated by Soft DCCA module. The correlation is computed based on the bi-temporal projected feature embeddings. Utilizing the proposed feature fusion module, the feature representation ability of both time could be enhanced. We also presented a new large-scale scene change detection dataset, and conducted experiments on this dataset. This dataset contains 23555 labeled scene image pairs with much more complex categories. Experimental results on this dataset and another smaller scene dataset both showed that our proposed CorrFusion module could remarkably improve the scene classification and scene change detection accuracies.
\par The rest sections of this paper are organized as follows. In Section \ref{dcca_sdcca}, we'll introduce some preliminary knowledge about DCCA and Soft DCCA. The detailed design, formulation and backpropagation derivation of proposed CorrFusion module will be presented in Section \ref{Methodology}. The introduction to our dataset and experiments are shown in Section \ref{Experiments}. In Section \ref{Conclusion}, the conclusions and potential outlooks of our work will be given.

\section{Related Works}
\label{dcca_sdcca}
\par Canonical Correlation Analysis (CCA) \cite{hotelling1992relations} is one of the most popular approaches in multi-view learning. In the remote sensing image processing field, CCA had also been widely applied and achieved pretty brilliant performances \cite{Nielsen2007, volpi2015spectral, yang2017multiview}.
\subsection{Deep Canonical Correlation Analysis}
\par In \cite{Andrew2013}, inspired by the success of DNN in representation learning, DCCA was proposed as an non-linear extension of CCA and proved to perform well in image recognition \cite{Yang2017}, cross-view feature extraction \cite{Rotman2018}, and image change detection \cite{yang2018heterogeneous, sahbi2018canonical}.
\par Assuming that $\mathbf{X}\in \mathbb{R}^{n\times d}$, $\mathbf{Y}\in \mathbb{R}^{n\times d}$ respectively denote the inputs of two views, where $n$ is the number of inputs and $d$  denote the number their dimensions. As shown is Fig~\ref{dcca}, DCCA firstly projects them into a new lower dimensional feature space with two independent DNNs. The outputs of the two branch DNNs are denoted by $\mathbf{X_\phi}=f(\mathbf{X}, \theta_1)\in \mathbb{R}^{n\times \frac{d}{r}}$ and $\mathbf{Y_\phi}=g(\mathbf{Y},\theta_2)\in \mathbb{R}^{n\times \frac{d}{r}}$, where $f$ and $g$ respectively denote the DNN projection function, $\theta_1$ and $\theta_2$ are correspondingly their parameters. The objective of DCCA is to maximize the sum of correlation between projected features, under the constraint that the projected features are both orthogonal. Formally, it's written as Eq.(\ref{dcca_tr}).
\begin{equation}
	\label{dcca_tr}
	\begin{split}
		argmax_{\theta_1, \theta_2}:\ tr(\mathbf{X_\phi}\mathbf{Y_\phi}^T), \\
		s.t.\ \mathbf{X_\phi}^T\mathbf{X_\phi}=\mathbf{Y_\phi}^T\mathbf{Y_\phi}=\mathbf{I},
	\end{split}
\end{equation}
where $\mathbf{I}$ denotes the identity matrix. The orthogonal constraint enforces the different dimensions of projected feature to be decorrelated. In \cite{Wang2016b}, an equivalent form of DCCA is presented as minimizing the $Frobenius\ Norm$ of the difference between $\mathbf{X}_\phi$ and $\mathbf{Y}_\phi$:
\begin{equation}
	\label{dcca_f}
	\begin{split}
		argmin_{\theta_1, \theta_2}:\ \frac{1}{2}||\mathbf{X_\phi}-\mathbf{Y_\phi}||_F, \\
		s.t.\ \mathbf{X_\phi}^T\mathbf{X_\phi}=\mathbf{Y_\phi}^T\mathbf{Y_\phi}=\mathbf{I}.
	\end{split}
\end{equation}
\par Since the orthogonal constraints are computed on all the training samples, the objectives and gradients could not be reliably estimated on a minibatch of samples when training a larger DCCA model on larger datasets \cite{Wang2016b,Wang2015}.
\subsection{Soft Deep Canonical Correlation Analysis}
\par
\begin{figure*}[]
	\centering
	\subfigure[DCCA] {
		\includegraphics[scale=0.42]{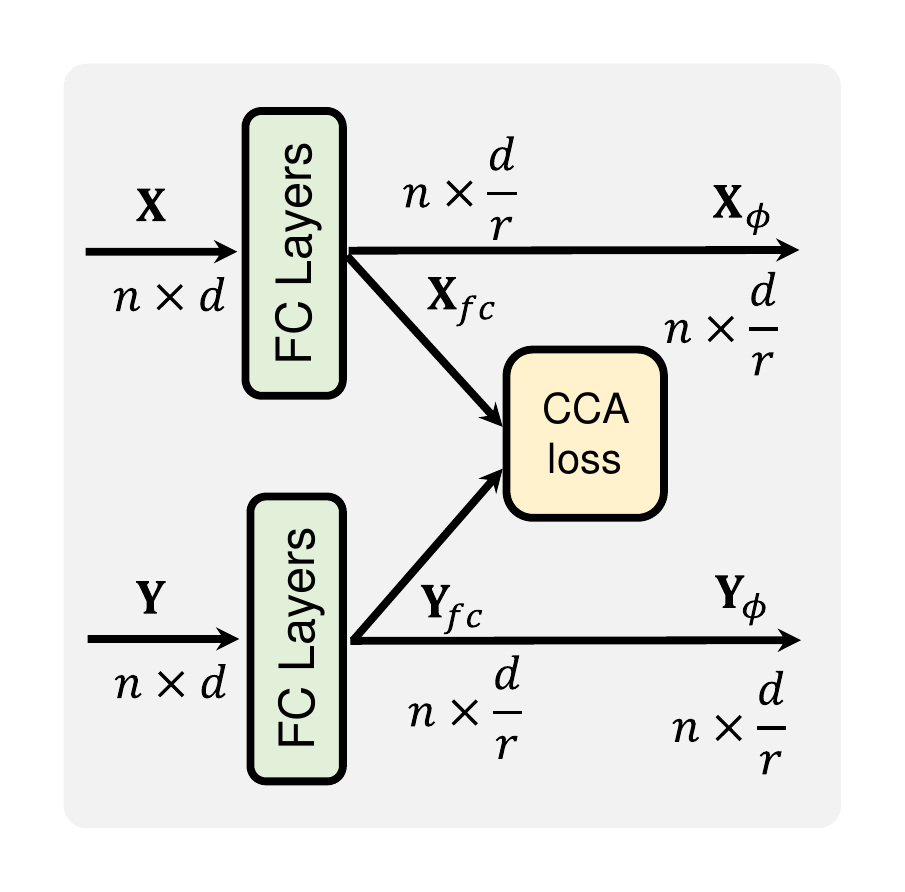}
		\label{dcca}
	}
	\subfigure[Soft DCCA] {
		\includegraphics[scale=0.42]{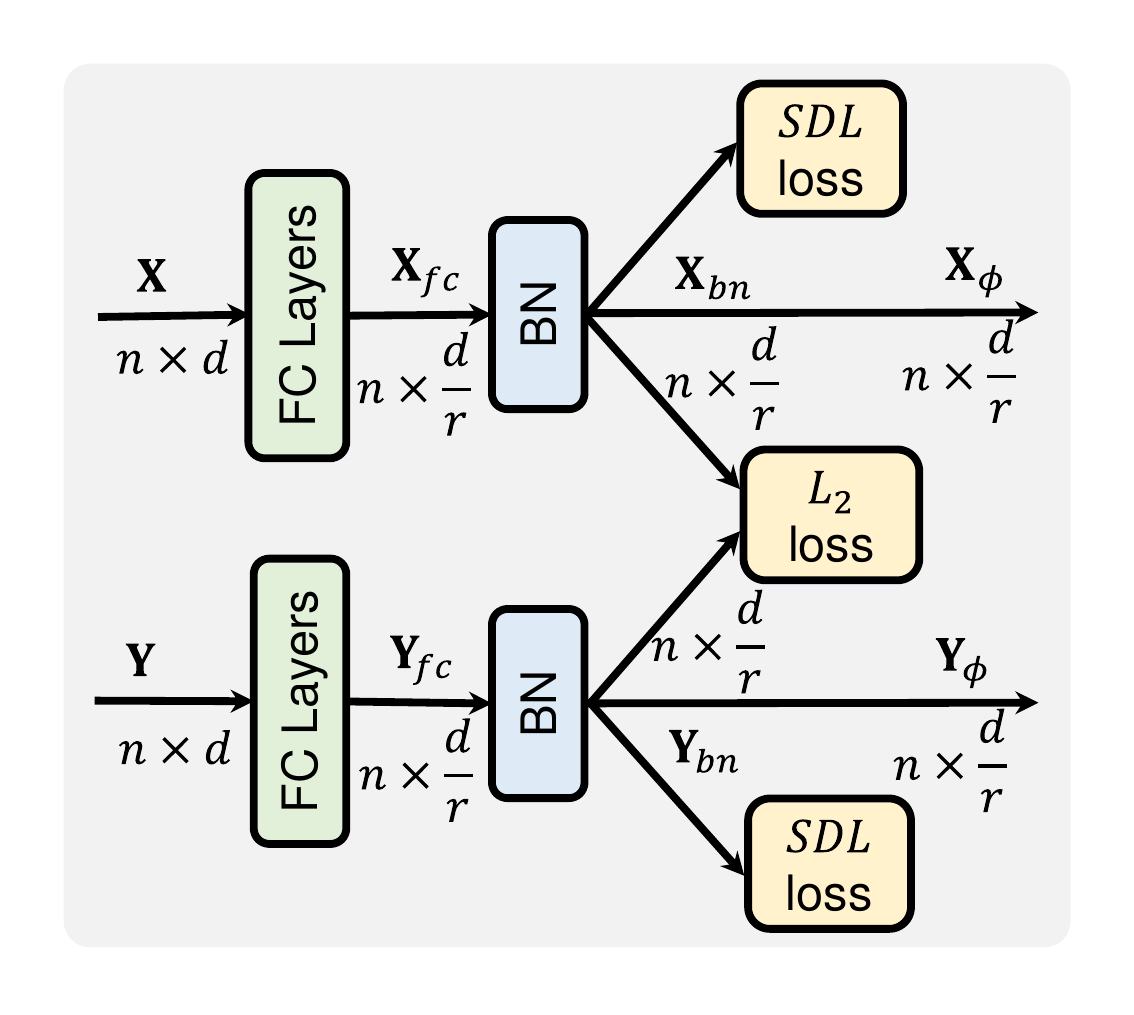}
		\label{sdcca}
	}
	\subfigure[CorrFusion] {
		\includegraphics[scale=0.42]{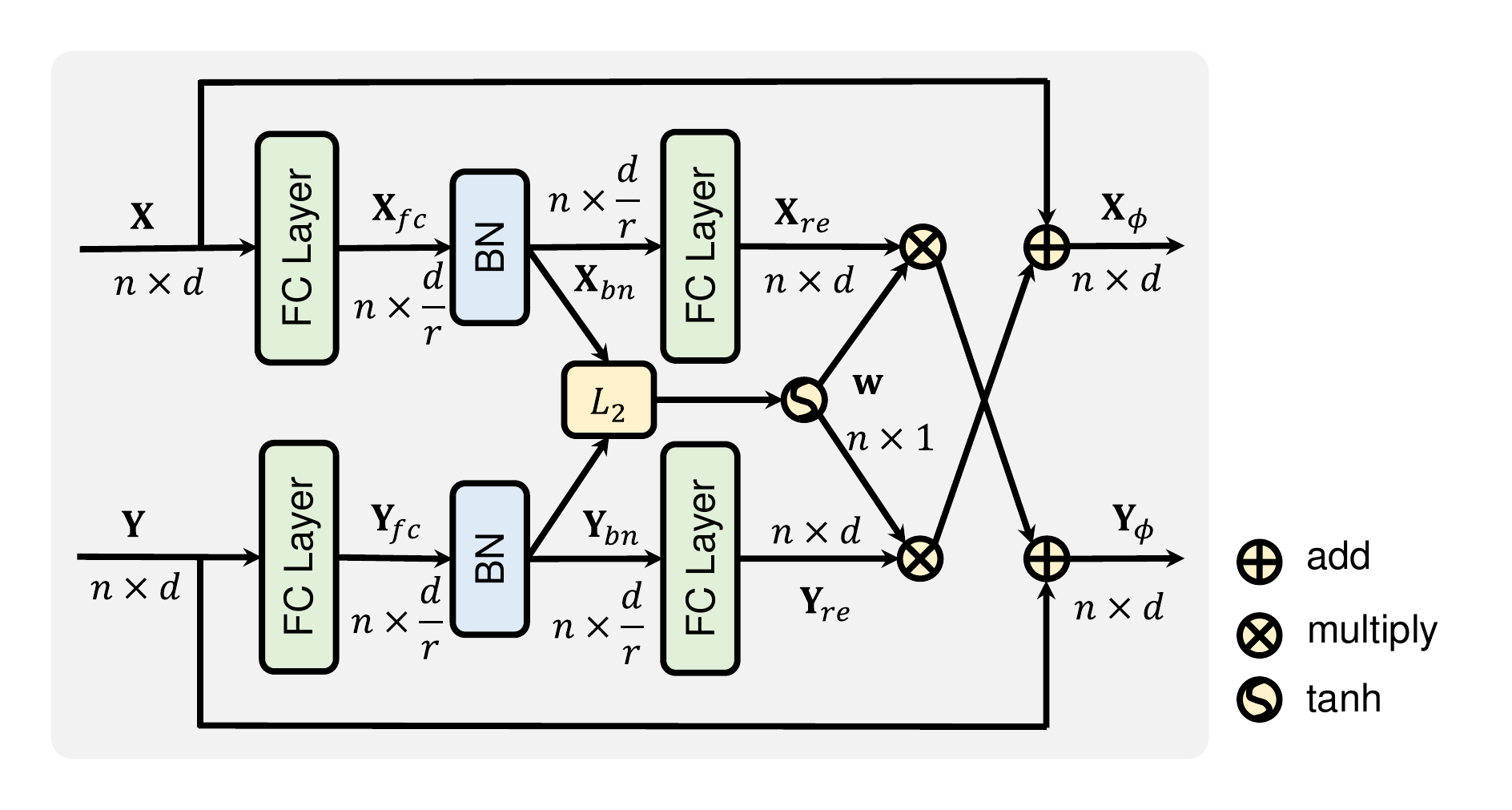}
		\label{CorrFusion}
	}

	\caption{The schematic diagrams of DCCA, Soft DCCA and our proposed CorrFusion module. $\mathbf{X}$ and $\mathbf{Y}$ are the bi-temporal inputs and $\mathbf{X}_\phi$ and $\mathbf{Y}_\phi$ respectively correspond to their outputs. $n$ and $d$ denote the number of the samples and feature dimensions, respectively. $r$ is a dimensionality-reduction ration. For simplicity, we left out the loss terms of CorrFusion in (c).}
	\label{fig_CorrFusion}
\end{figure*}

 To solve this problem, based on the formulation in Eq.(\ref{dcca_f}), \textit{Chang et al.} proposed Soft DCCA \cite{Chang2018a}. As presented in Fig~\ref{sdcca}, the key idea of Soft DCCA is to relax the original hard orthogonal constraints with $SDL$ loss \cite{Cogswell2015}.
\par Given the $i$-th minibatch projected embeddings $\mathbf{X}^i_\phi\in\mathbb{R}^{n_i\times d}$ with $n_i$ is the number of samples in this batch. We could further assume that $\mathbf{X}^i_\phi$ is centralized, which could be easily accomplished with a batch normalization layer \cite{Ioffe2015}. Soft DCCA firstly computes the covariance matrix of the $i$-th batch as
\begin{equation}
	\mathbf{\Sigma}^i_{\mathbf{XX}} = \frac{1}{n_i-1}{\mathbf{X}^i_\phi}^T{\mathbf{X}^i_\phi}.
\end{equation}
\par As aforementioned, the covariance matrix could not be reliably estimated on a minibatch. Following the solution in \cite{Wang2016b}, Soft DCCA computes the estimation with an accumulative mechanism.
\begin{equation}
	\label{stable_cov}
	\begin{split}
		\mathbf{\tilde{\Sigma}}^i_{\mathbf{XX}} &= \rho\mathbf{\tilde{\Sigma}}^{i-1}_{\mathbf{XX}}+(1-\rho)\frac{1}{n_i-1}{\mathbf{X}^i_\phi}^T{\mathbf{X}^i_\phi},\\
		\mathbf{\tilde{\Sigma}}^0_{\mathbf{XX}} &= \frac{1}{n_0-1}{\mathbf{X}^0_\phi}^T{\mathbf{X}^0_\phi},
	\end{split}
\end{equation}
where $\rho\in[0,1)$ is a momentum parameter, and $\mathbf{\tilde{\Sigma}}^0_{\mathbf{XX}}$ is the initial covariance matrix computed on a random batch. Soft DCCA then replaces the hard orthogonal constraint in Eq.(\ref{dcca_f}) with softer decorrelation loss by minimizing the sum absolute value of the off-diagonal entries of $\mathbf{\tilde{\Sigma}}^i_{\mathbf{XX}}$:
\begin{equation}
	\mathbf{\mathcal{L}}_{SDL}(\mathbf{X}^i_\phi)=\sum_{k=1}^{d}\sum_{l=1,l\neq k}^{d}|\mathbf{\tilde{\Sigma}}^i_{\mathbf{XX}}|_{kl}.
\end{equation}
By minimizing the $SDL$ loss of $\mathbf{X}_\phi$ and $\mathbf{Y}_\phi$, Soft DCCA is then defined as an unconstrained optimization problem:
\begin{equation}
	\label{Soft DCCA}
	min:\mathbf{\mathcal{L}}_{2}(\mathbf{X_\phi},\mathbf{Y_\phi})+\mathbf{\mathcal{L}}_{SDL}(\mathbf{X}_\phi)+\mathbf{\mathcal{L}}_{SDL}(\mathbf{Y}_\phi),
\end{equation}
with $\mathbf{\mathcal{L}}_{2}(\mathbf{X_\phi},\mathbf{Y_\phi})$ denotes the $\mathbf{\mathcal{L}}_{2}$ distance between $\mathbf{X}_\phi$ and $\mathbf{Y}_\phi$. All the terms in Eq.(\ref{Soft DCCA}) could be stably minimized with minibatch gradient descent optimizers (e.g. SGD). An intuitive interpretation of Soft DCCA is that as $\mathbf{\mathcal{L}}_{SDL}(\mathbf{X}_\phi)\to 0$, $\mathbf{X}_\phi$ approaches an orthogonal matrix, so that the constraints in Eq.(\ref{dcca_f}) is satisfied. Besides, since the $\mathbf{\mathcal{L}}_{2}$ distance between $\mathbf{X}_\phi$ and $\mathbf{Y}_\phi$ is equivalent to the objective function in Eq.(\ref{dcca_f}), the objective in Eq.(\ref{dcca_tr}) could finally be maximized by minimizing Eq.(\ref{Soft DCCA}).

\section{Methodology}
\label{Methodology}

\par In this section, we'll firstly amplify the detailed design and formulation of our proposed CorrFusion module, which is the key part to perform temporal correlation computation and feature fusion. Then an multi-temporal scene classification network with the proposed CorrFusion will be introduced. Besides, the derivation of the backpropagation gradients of CorrFusion module will also be presented.

\subsection{CorrFusion Module}

\par As presented in Fig~\ref{CorrFusion}, the CorrFusion module takes the bi-temporal features $\mathbf{X}$ and $\mathbf{Y}$ as inputs. Then a fully connected layer and a batch normalization layer are respectively employed to project $\mathbf{X}$ and $\mathbf{Y}$ into a lower dimension feature space and normalize the features. Based on the normalized features $\mathbf{X}_{bn}$ and $\mathbf{Y}_{bn}$, the instance-level temporal correlation is calculated as the $\mathcal{L}_2$ distance between the features of each scene image pair. The weight vector $\mathbf{w}$ is then computed by scaling the temporal correlation to $(0,1)$ using $\tanh$ function. Next, $\mathbf{X}_{bn}$ and $\mathbf{Y}_{bn}$ will be restored to the same dimension with $\mathbf{X}$ and $\mathbf{Y}$ by a dimensionality-increasing layer, which is also a fully connected layer actually. Since the weight vector $\mathbf{w}$ modeled the similarity between bi-temporal scene image pairs, a cross-temporal addition operation between bi-temporal features with high similarity will improve the reliability of feature representation.
\par To be specific, mathematically, let's assume that the bi-temporal input deep features are respectively $\mathbf{X}^i\in\mathbb{R}^{n_i\times d}$ and $\mathbf{Y}^i\in\mathbb{R}^{n_i\times d}$ in the $i$-th batch, where $n_i$ is the size of this batch. For $\mathbf{X}^i$, we firstly project it into a lower dimension feature space by a dimensionality-reduction layer. The weight matrix and bias vector are respectively $\mathbf{W}_{fc}\in \mathbb{R}^{d\times \frac{d}{r}}$ and $\mathbf{b}_{fc}\in \mathbb{R}^{\frac{d}{r}}$, with $r$ is a reduction ratio. The output of the dimensionality-reduction layer is computed as
\begin{equation}
	\mathbf{X}^i_{fc} = s(\mathbf{X}^i\mathbf{W}_{fc}+\mathbf{b}_{fc}),
	\label{x_fc}
\end{equation}
where $s(\cdot)$ denotes the activation function. $\mathbf{X}^i_{fc}$ is then passed through a batch normalization layer to implement the normalization constraint intrinsically indicated in Eq.(\ref{dcca_f}).
\begin{equation}
	\label{x_bn}
	\mathbf{X}^i_{bn} = batch\_norm(\mathbf{X}^i_{fc}),
\end{equation}
where $batch\_norm(\cdot)$ denotes the transformation function in a batch normalization layer. The calculation and expression of $\mathbf{Y}^i_{bn}$ are symmetric. Following Eq.(\ref{stable_cov}), we maintain accumulative estimations of the covariance matrix for $\mathbf{X}^i_{bn}$ and $\mathbf{Y}^i_{bn}$ respectively as
\begin{equation}
	\label{xy_cov}
	\begin{split}
		\mathbf{\tilde{\Sigma}}^i_{\mathbf{XX}} &= \rho\mathbf{\tilde{\Sigma}}^{i-1}_{\mathbf{XX}}+(1-\rho)\frac{1}{n_i-1}{\mathbf{X}^i_{bn}}^T{\mathbf{X}^i_{bn}},\\
		\mathbf{\tilde{\Sigma}}^i_{\mathbf{YY}} &= \rho\mathbf{\tilde{\Sigma}}^{i-1}_{\mathbf{YY}}+(1-\rho)\frac{1}{n_i-1}{\mathbf{Y}^i_{bn}}^T{\mathbf{Y}^i_{bn}},
	\end{split}
\end{equation}
with $\rho\in[0,1)$ is still the momentum parameter. Their $SDL$ loss are calculated as
\begin{equation}
	\label{xy_sdl}
	\begin{split}
		\mathbf{\mathcal{L}}_{SDL}(\mathbf{X}^i_{bn})=\sum_{k=1}^{d/r}\sum_{l=1,l\neq k}^{d/r}|\mathbf{\tilde{\Sigma}}^i_{\mathbf{XX}}|_{kl},\\
		\mathbf{\mathcal{L}}_{SDL}(\mathbf{Y}^i_{bn})=\sum_{k=1}^{d/r}\sum_{l=1,l\neq k}^{d/r}|\mathbf{\tilde{\Sigma}}^i_{\mathbf{YY}}|_{kl}.
	\end{split}
\end{equation}
\par By minimizing $\mathbf{\mathcal{L}}_{SDL}(\mathbf{X}^i_{bn})$ and $\mathbf{\mathcal{L}}_{SDL}(\mathbf{Y}^i_{bn})$, $\mathbf{X}^i_{bn}$ and $\mathbf{Y}^i_{bn}$ approach orthogonal matrices, which also ensures that they won't be $\textbf{0}$. The instance level correlation between $\mathbf{X}^i_{bn}$ and $\mathbf{Y}^i_{bn}$ is defined as the $\mathcal{L}_2$ norm of each row in their difference matrix.
\begin{equation}
	\label{xy_l}
	\ell(k)=||\mathbf{X}^i_{bn}(k,:)-\mathbf{Y}^i_{bn}(k,:)||_2, k=(1,2,\cdots,n_i),
\end{equation}
which can be converted to the objective function in Eq.(\ref{dcca_f}) by
\begin{equation}
	\label{xy_f}
	||\mathbf{X}^i_{bn}-\mathbf{Y}^i_{bn}||_F=({\sum_{k=1}^{n_i}\ell(k)^2})^{\frac{1}{2}}.
\end{equation}
\par In Eq.(\ref{xy_l}), each entry of $\ell$ denotes the distance between the corresponding sample pairs from $\mathbf{X}^i_{bn}$ and $\mathbf{Y}^i_{bn}$. To obtain the weight vector $\mathbf{w}$, we then scale $\ell$ to $(0,1)$ with $\tanh$ function. Besides, $\mathbf{w}$ should be monotone and also satisfy that $\mathbf{w}\rightarrow 1$ when $\ell\rightarrow 0$.
\begin{equation}
	\label{xy_w}
	\mathbf{w}=1-\tanh(\ell).
\end{equation}
\par Based on Eq.(\ref{x_bn}), we could perform dimensionality-increasing with a fully connected layer. Assuming that the weight matrix and bias vector are respectively $\mathbf{W}_{re}\in \mathbb{R}^{\frac{d}{r}\times d}$ and $\mathbf{b}_{re}\in \mathbb{R}^{d}$ in this layer, the restored $\mathbf{X}^i_{re}$ from $\mathbf{X}^i_{bn}$ is formulated as
\begin{equation}
	\label{x_re}
	\mathbf{X}^i_{re}= s(\mathbf{X}^i_{bn}\mathbf{W}_{re}+\mathbf{b}_{re}).
\end{equation}
\par Considering that we have calculated the temporal correlation in $\mathbf{w}$, a larger $\mathbf{w}(k)$ indicates the corresponding $\mathbf{X}^i(k,:)$ and $\mathbf{Y}^i(k,:)$ are more likely to be sampled from the same scene category. Therefore, it's a natural idea to boost the feature representation ability by adding the weighted embeddings from the other branch.
\begin{equation}
	\label{x_add}
	\mathbf{X}^i_{\phi}(k,:)= \mathbf{X}^i(k,:)+\mathbf{w}(k)\mathbf{Y}^i_{re}(k,:).
\end{equation}
\par With a cross-temporal addition operation in Eq.(\ref{x_add}), the dimensionality-increasing result of the $k$-th sample of $\mathbf{Y}^i$, which is highly correlated with $\mathbf{X}^i(k,:)$, will be added to the original inputs $\mathbf{X}$ with a large weight. On the contrary, $\mathbf{Y}^i(k,:)$ with lower correlation with $\mathbf{X}^i(k,:)$ will get a small $\mathbf{w}(k)$ in the consequent calculation, thus won't impact much on the embedding distribution of $\mathbf{X}^i(k,:)$. $\mathbf{Y}^i_{\phi}(k,:)$ has a dual expression with $\mathbf{X}^i_{\phi}(k,:)$ in Eq.(\ref{x_add}), which is formulated as
\begin{equation}
	\label{y_add}
	\mathbf{Y}^i_{\phi}(k,:)= \mathbf{Y}^i(k,:)+\mathbf{w}(k)\mathbf{X}^i_{re}(k,:).
\end{equation}
\par In Algorithm \ref{alg-corrfusion}, we provide the pseudocode of an implementation for the proposed CorrFusion module in TensorFlow-style. The implementations of computing the accumulative covariance matrix, $SDL$ loss and bi-temporal correlation are also included.

\begin{algorithm}
	\caption{Pseudocode in TensorFlow-style for the proposed CorrFusion module.}
	\label{alg-corrfusion}
	\definecolor{codeblue}{rgb}{0.25,0.25,0.6}
	\definecolor{codegreen}{rgb}{0.25,0.6,0.25}
	\lstset{
		backgroundcolor=\color{white},
		basicstyle=\fontsize{7.7pt}{7.7pt}\ttfamily\selectfont,
		columns=fullflexible,
		breaklines=true,
		captionpos=b,
		commentstyle=\fontsize{7.7pt}{7.7pt}\color{codegreen},
		keywordstyle=\fontsize{7.7pt}{7.7pt}\color{codeblue},
	}
	\begin{lstlisting}[language=python]
def CorrFusion(x=None, y=None):
    N = tf.shape(input=x)[0]
    ## dimensionality reduction
    x_fc = fc_layer(x, units=dim_r)
    y_fc = fc_layer(y, units=dim_r)
    ## batch normalization
    x_bn = bn_layer(x_fc, axis=-1)
    y_bn = bn_layer(y_fc, axis=-1)
    ## dimensionality increasing
    x_re = fc_layer(x_bn, units=dim)
    y_re = fc_layer(y_bn, units=dim)
    ## compute the instance-level correlation
    corr_s = tf.reduce_sum(tf.square(x_bn - y_bn))
    corr = tf.sqrt(corr_s, axis=-1)
    ## compute the accumulative covariance
    x_cov = rho*x_cov + (1-rho)*tf.matmul(x_bn, x_bn, transpose_a=True) / (N-1)
    y_cov = rho*y_cov + (1-rho)*tf.matmul(y_bn, y_bn, transpose_a=True) / (N-1)
    ## compute the decorrelation loss
    with tf.name_scope('decorrelation'):
        x_SDL = tf.reduce_sum(tf.abs(x_cov)) - tf.reduce_sum(tf.diag_part(x_cov))
        y_SDL = tf.reduce_sum(tf.abs(y_cov)) - tf.reduce_sum(tf.diag_part(y_cov))
        SDL_loss = tf.reduce_mean(x_SDL + y_SDL)
    ## cross temporal fusion
    with tf.name_scope('fusion'):
        w = 1-tf.nn.tanh(tf.expand_dims(corr, axis=1))
        wx = tf.multiply(x_re, w)
        wy = tf.multiply(y_re, w)
    x_phi = x + wy; y_phi = y + wx
    return x_phi, y_phi, SDL_loss
\end{lstlisting}
\end{algorithm}

\subsection{Network Overview}
\begin{figure}[]
	\centering
	\includegraphics[scale=0.45]{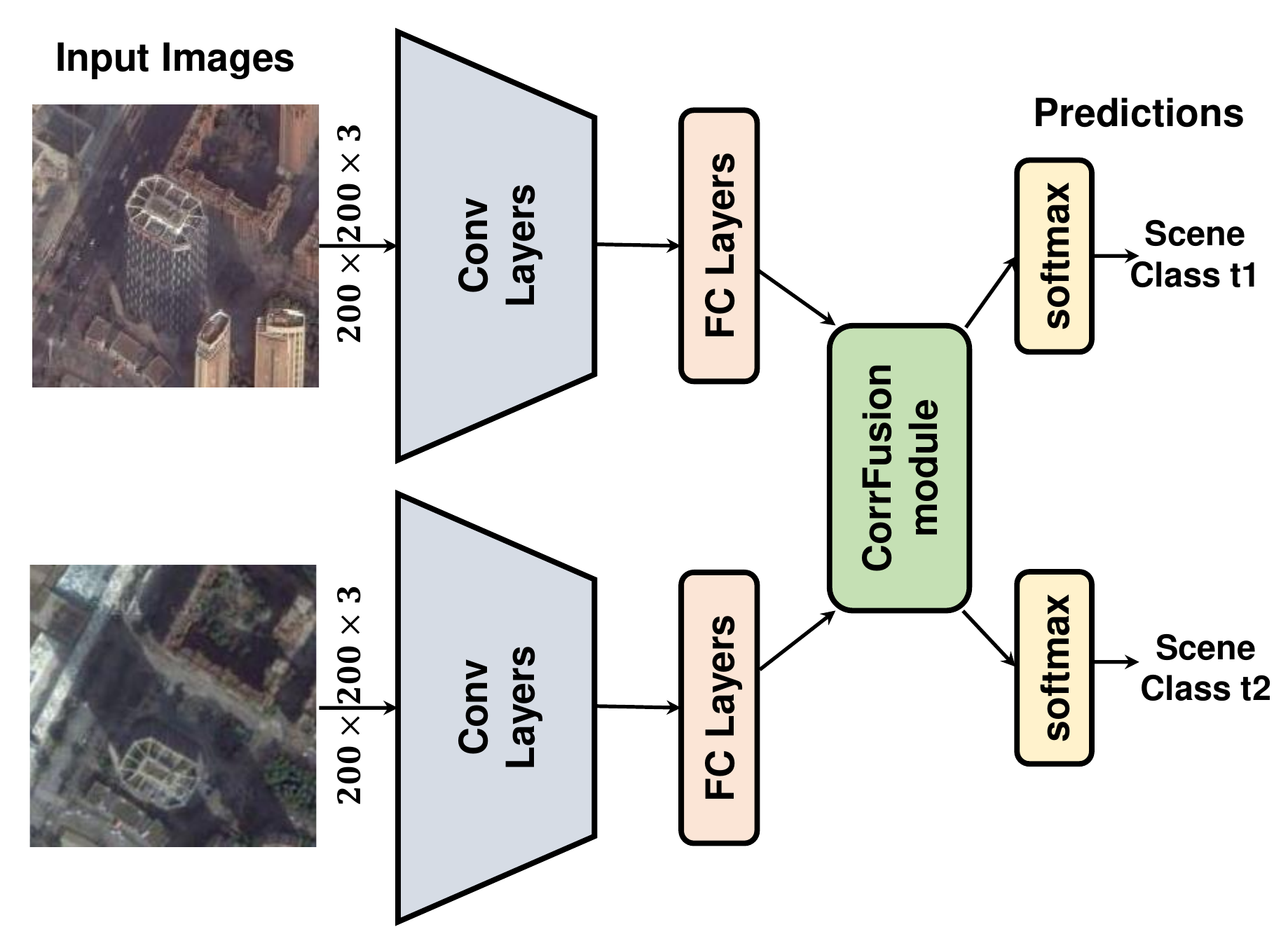}
	\caption{The proposed multi-temporal scene classification and scene change detection framework with CorrFusion module. It contains two branch convolutional module to extract the deep representations of bi-temporal input scene images. The extracted features are then projected into a new feature space by multiple fully connected layers. Then the CorrFusion module is employed to compute the correlation between the bi-temporal features and perform feature fusion to enhance the feature representation capacity. The scene categories are finally obtained with softmax layers.}
	\label{fig_CorrFusionNet}
\end{figure}
\par Based on the CorrFusion module, we present an end-to-end scene change detection framework called CorrFusionNet in Fig~\ref{fig_CorrFusionNet}. As shown in Fig~\ref{fig_CorrFusionNet}, CorrFusionNet utilizes two branch deep convolutional network to perform deep feature extraction for the bi-temporal input images. The extracted features will be projected into a new embedding space by fully connected layers.
\par Formally, assuming the $i$-th batch of the input images for the branch 1 is $\mathbf{I}^i_\mathbf{x}$, and the output of the last fully connected layer for is denoted as $\mathbf{X}^i$, which is exactly the $\mathbf{X}^i$ in Eq.(\ref{x_fc}). For another branch, we could obtain $\mathbf{Y}^i$ for $\mathbf{I}^i_\mathbf{y}$ in a similar way.
\par After obtaining the bi-temporal inputs for the aforementioned CorrFusion module, $\mathbf{X}^i_\phi$ and $\mathbf{Y}^i_\phi$ could then be computed following the formulations from Eq.(\ref{x_fc}) to Eq.(\ref{x_add}). In subsequent, a softmax activation layer is employed to calculate the predicted probability for scene classification. Let's assume the predictions for $\mathbf{I}^i_\mathbf{x}$ and $\mathbf{I}^i_\mathbf{y}$ are respectively $p^i_\mathbf{x}$ and $p^i_\mathbf{x}$, and their corresponding true labels are $l^i_\mathbf{x}$ and $l^i_\mathbf{y}$, the final loss function of proposed CorrFusionNet is formulated as
\begin{equation}
	\label{loss}
	\begin{split}
		\mathbf{\mathcal{L}}_{total}&=\mathbf{\mathcal{L}}_{CE}(p^i_\mathbf{x},l^i_\mathbf{x})+\mathbf{\mathcal{L}}_{CE}(p^i_\mathbf{y},l^i_\mathbf{y})\\
		&+\mathcal{L}_{corr}(\mathbf{X}^i_{bn},\mathbf{Y}^i_{bn};l^i_\mathbf{x}, l^i_\mathbf{x})\\
		&+(\mathbf{\mathcal{L}}_{SDL}(\mathbf{X}^i_{bn})+\mathbf{\mathcal{L}}_{SDL}(\mathbf{Y}^i_{bn})).
	\end{split}
\end{equation}

\par In Eq.(\ref{loss}), $\mathcal{L}_{CE}(p^i_\mathbf{x},l^i_\mathbf{x})$ denotes the cross entropy loss computed on the predicted and true labels of $\mathbf{I}^i_\mathbf{x}$. The third term denotes the $SDL$ constraints calculated on $\mathbf{X}^i_{bn}$ and $\mathbf{Y}^i_{bn}$. The second term denotes the objective in Eq.(\ref{dcca_f}).Besides, since bi-temporal scene images don't always belong to the same scene category, we utilize $\xi$ to only compute the correlation between the unchanged scene image pairs. $\xi$ is required to satisfy $\xi(k)=1$ if $l^i_\mathbf{x}(k)=l^i_\mathbf{y}(k)$, and $\xi(k)=0$ otherwise. Based on Eq.(\ref{xy_l}) and Eq.(\ref{xy_f}), $\mathcal{L}_{corr}(\mathbf{X}^i_{bn},\mathbf{Y}^i_{bn};l^i_\mathbf{x}, l^i_\mathbf{x})$ is written as:
\begin{equation}
	\label{l_corr}
	\begin{split}
		\mathcal{L}_{corr}(\mathbf{X}^i_{bn},\mathbf{Y}^i_{bn};l^i_\mathbf{x}, l^i_\mathbf{x})
		&=({\sum_{k=1}^{n_i}\xi(k)\ell(k)^2})^{\frac{1}{2}}\\
		&=({\sum_{k=1}^{n_i}\xi(k)||\mathbf{X}^i_{bn}(k,:)-\mathbf{Y}^i_{bn}(k,:)||_2^2})^{\frac{1}{2}}.
	\end{split}
\end{equation}
\par It's noted that all terms in Eq.(\ref{loss}) are all minibatch based losses, which indicates they could be stably estimated and optimized by minibatch gradient descent algorithms. Besides, our proposed CorrFusionNet maintains an end-to-end structure, so that it could be trained in a whole without multi-stage processing.
\subsection{Optimization}
\par We will present how the gradient is backpropagated over the CorrFusion module in this section. Let's firstly consider the entry in the $k$-th row and $l$-th column in the outputs $\mathbf{X}_\phi$:
\begin{equation}
	\mathbf{X}_{\phi}^{kl}= \mathbf{X}^{kl}+\mathbf{w}(k)\mathbf{Y}_{re}^{kl}.
\end{equation}
\par Its gradient with respect to $\mathbf{X}$ and $\mathbf{Y}$ are respectively defined as
\begin{equation}
	\label{x_grad}
	\begin{split}
		\frac{\partial{\mathbf{X}_\phi^{kl}}}{\partial{\mathbf{X}}}&=\frac{\partial{\mathbf{X}^{kl}}}{\partial{\mathbf{X}}}+\frac{\partial{\mathbf{w}(k)}}{\partial{\mathbf{X}}}\mathbf{Y}_{re}^{kl},\\
		\frac{\partial{\mathbf{X}_\phi^{kl}}}{\partial{\mathbf{Y}}}&=\frac{\partial{\mathbf{w}(k)}}{\partial{\mathbf{Y}}}\mathbf{Y}_{re}^{kl}+\mathbf{w}(k)\frac{\partial{\mathbf{Y}_{re}^{kl}}}{\partial{\mathbf{Y}}}.
	\end{split}
\end{equation}
\par The result for $\partial{\mathbf{X}^{kl}}/\partial{\mathbf{X}}$ is obvious. Besides, since the backpropagation from $\mathbf{X}_{bn}$ to $\mathbf{X}$ only involves the gradients on fully connected and BN layers, we could only compute  $\partial{\mathbf{X}_\phi^{kl}}/\partial{\mathbf{X}_{bn}}$ and $\partial{\mathbf{X}_\phi^{kl}}/\partial{\mathbf{Y}_{bn}}$ and then use chain rule. Therefore, the less trivial part of derivation in Eq.(\ref{x_grad}) is simplified as
\begin{equation}
	\label{xbn_grad}
	\begin{split}
		\frac{\partial{\mathbf{X}_\phi^{kl}}}{\partial{\mathbf{X}_{bn}}}&=\frac{\partial{\mathbf{w}(k)}}{\partial{\mathbf{X}_{bn}}}\mathbf{Y}_{re}^{kl},\\
		\frac{\partial{\mathbf{X}_\phi^{kl}}}{\partial{\mathbf{Y}_{bn}}}&=\frac{\partial{\mathbf{w}(k)}}{\partial{\mathbf{Y}_{bn}}}\mathbf{Y}_{re}^{kl}+\mathbf{w}(k)\frac{\partial{\mathbf{Y}_{re}^{kl}}}{\partial{\mathbf{Y}_{bn}}},
	\end{split}
\end{equation}
\par Integrating Eq.(\ref{xy_w}), for the gradient of $\mathbf{w}(k)$ w.r.t $\mathbf{X}_{bn}$, we have
\begin{equation}
	\label{w_grad}
	\frac{\partial{\mathbf{w}(k)}}{\partial{\mathbf{X}_{bn}}}=-\frac{\partial{\tanh{\ell(k)}}}{\partial{\mathbf{X}_{bn}}}=(\tanh^2\ell(k)-1)\frac{\partial\ell(k)}{\partial{\mathbf{X}_{bn}}}.
\end{equation}
\par Based on the definition of $\ell(k)$ in Eq.(\ref{xy_l}) and derivations in \cite{matrixcookbook}, we should have
\begin{equation}
	\label{l_grad}
	{\frac{\partial\ell(k)}{\partial{\mathbf{X}_{bn}}}}=\frac{\mathbf{I}^{k}}{\ell(k)}\odot(\mathbf{X}_{bn}-\mathbf{Y}_{bn}).
\end{equation}
\par In Eq.(\ref{l_grad}), $\mathbf{I}^{k}$ denotes a matrix with only entries in the $k$-th row are 1, and $\odot$ is the dot multiplication operation between matrices. By integrating Eq.(\ref{w_grad}) to Eq.(\ref{l_grad}), $\partial{\mathbf{w}(k)}/{\partial{\mathbf{X}_{bn}}}$ is shown as
\begin{equation}
	\label{w_grad_x}
	\frac{\partial{\mathbf{w}(k)}}{\partial{\mathbf{X}_{bn}}}=\frac{\tanh^2\ell(k)-1}{\ell(k)}\mathbf{I}^{k}\odot(\mathbf{X}_{bn}-\mathbf{Y}_{bn}).
\end{equation}
\par The expression for ${\partial{\mathbf{w}(k)}}/{\partial{\mathbf{Y}_{bn}}}$ is symmetric. We could then rewrite Eq.(\ref{xbn_grad}) as
\begin{equation}
	\label{xy_grad}
	\begin{split}
		\frac{\partial{\mathbf{X}_\phi^{kl}}}{\partial{\mathbf{X}_{bn}}}&=\mathbf{Y}_{re}^{kl}\frac{\tanh^2\ell(k)-1}{\ell(k)}\mathbf{I}^{k}\odot(\mathbf{X}_{bn}-\mathbf{Y}_{bn}),\\
		\frac{\partial{\mathbf{X}_\phi^{kl}}}{\partial{\mathbf{Y}_{bn}}}&=\mathbf{Y}_{re}^{kl}\frac{\tanh^2\ell(k)-1}{\ell(k)}\mathbf{I}^{k}\odot(\mathbf{Y}_{bn}-\mathbf{X}_{bn})+\mathbf{w}(k)\mathbf{\Delta},
	\end{split}
\end{equation}
with $\mathbf{\Delta}$ is the gradient across the dimensionality-increasing layer (a fully connected layer) and is supposed to satisfy $\mathbf{\Delta}^{ij}=\mathbf{W}_{re}^{ji},\ if\ i=k;\ \mathbf{\Delta}^{ij}=0,\ otherwise$.
\par In terms of the computational complexity, except for the dimensionality manipulation layers in Eq.(\ref{x_fc}) and Eq.(\ref{x_re}), the matrix multiplication, matrix addition and element-wise multiplication are also involved in the proposed CorrFusion module. The computational complexity of them are respectively $\mathcal{O}(nd^2)$, $\mathcal{O}(d^2)$, and $\mathcal{O}(d^2)$. Therefore, the complexity for the proposed module is $\mathcal{O}(nd^2)$, and is lower than the complexity of the formulations of DCCA in \cite{Andrew2013, Wang2016b}, which use SVD to perform exact decorrelation and have a complexity of $\mathcal{O}(d^3+nd^2)$.
\section{Experiments}
\label{Experiments}

\subsection{Dataset Description}

\subsubsection{{Hanyang Dataset}}
Hanyang Dataset is a small open-accessed scene change detection dataset. It only contains 190 training image pairs and 1050 testing image pairs with 8 land use scene categories. The spatial resolution and size of these images are 1m and $150\times150$, respectively.

\subsubsection{Wuhan Dataset}
\par Existing scene change detection datasets used in \cite{Wu2016,Wu2017} are all small in scale and only contain very few number of land-use scene categories. In this work, we labeled a much larger dataset which contains more scene categories.
\label{wuhan_description}
\begin{figure*}[h!t] \centering
	\subfigure[] {
		\includegraphics[width=85mm]{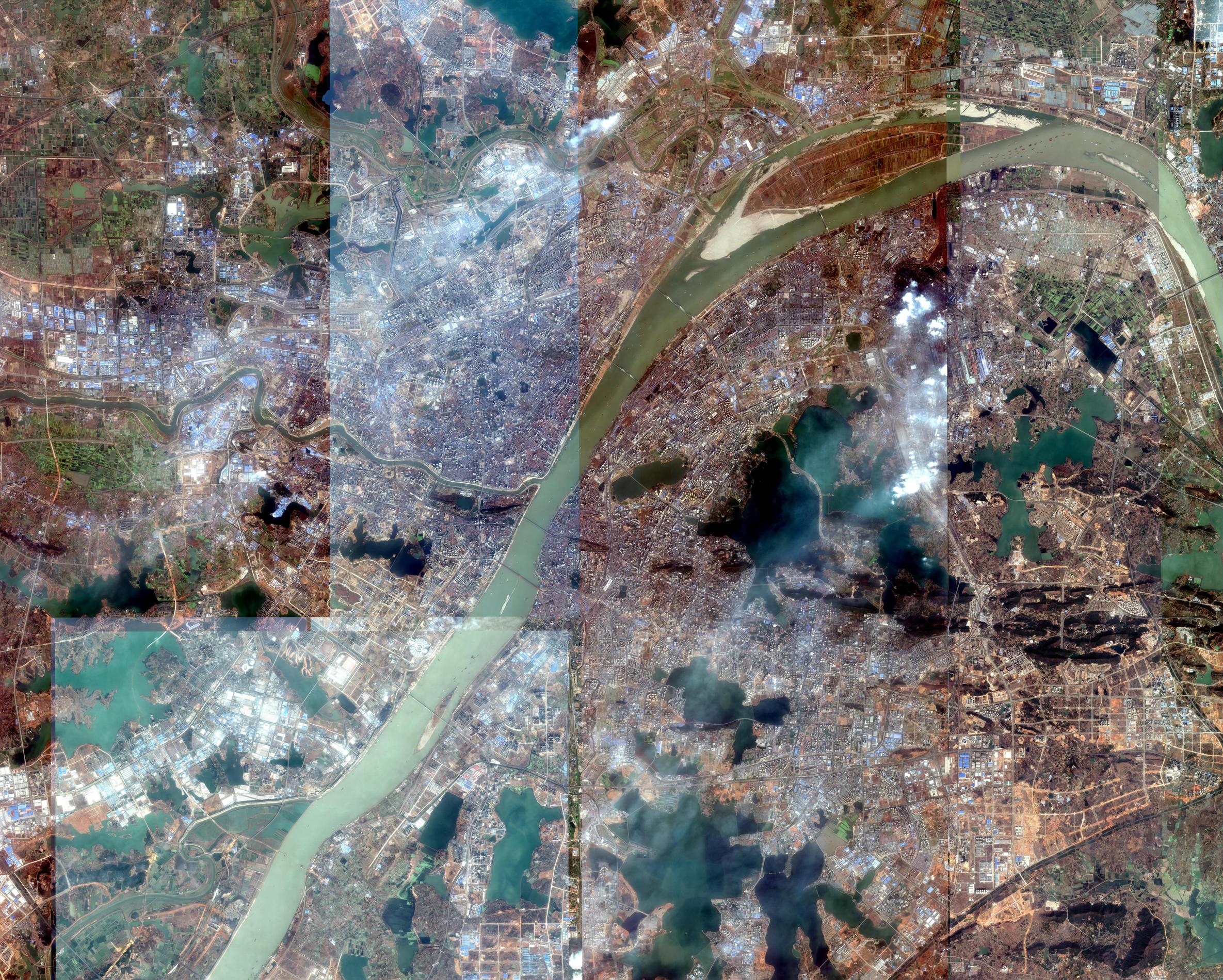}
		\label{wuhan_2014}
	}
	\hspace{-0.15in}
	\subfigure[] {
		\includegraphics[width=85mm]{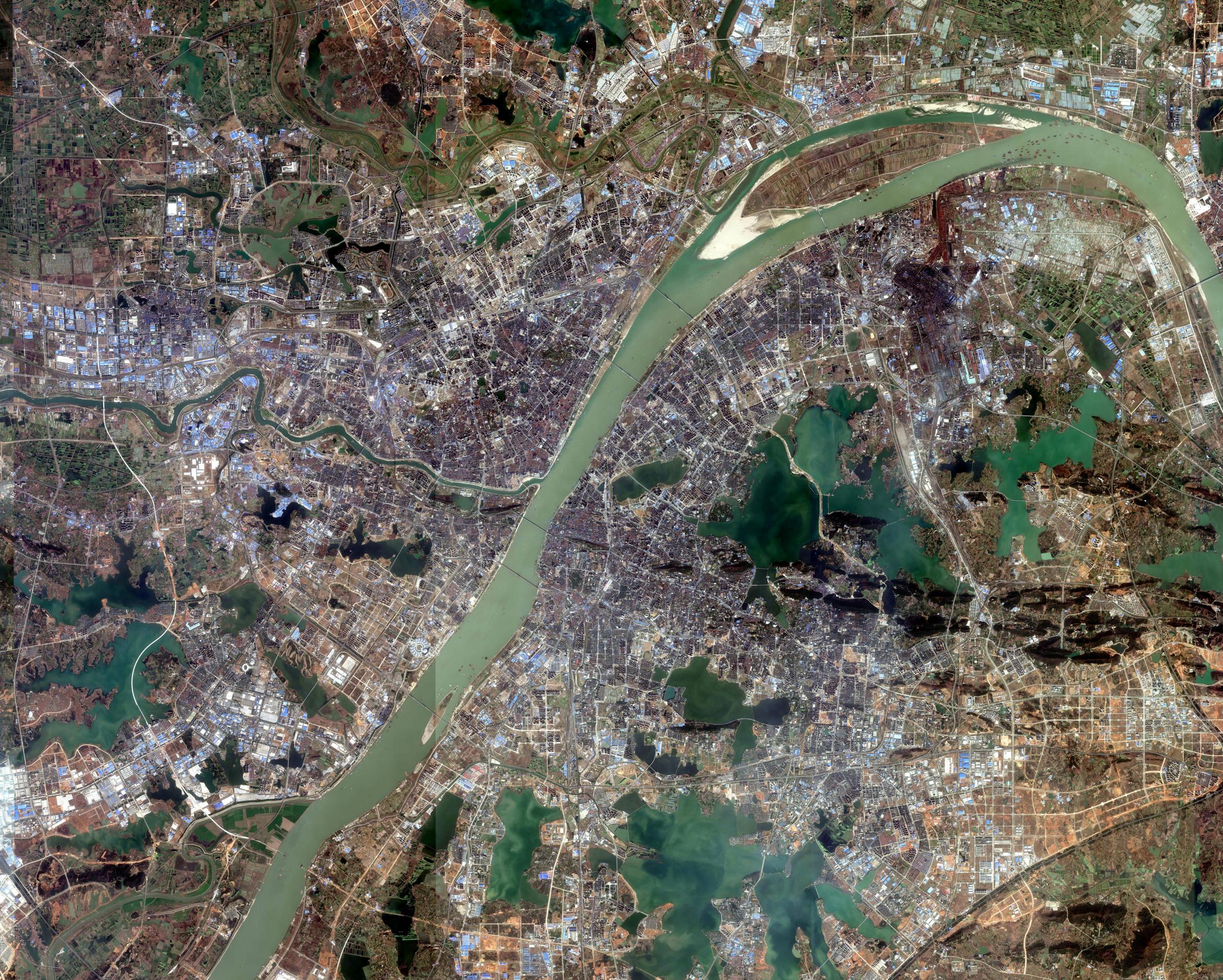}
		\label{wuhan_2016}
	}
	\hspace{-0.15in}

	\subfigure[] {
		\includegraphics[width=85mm]{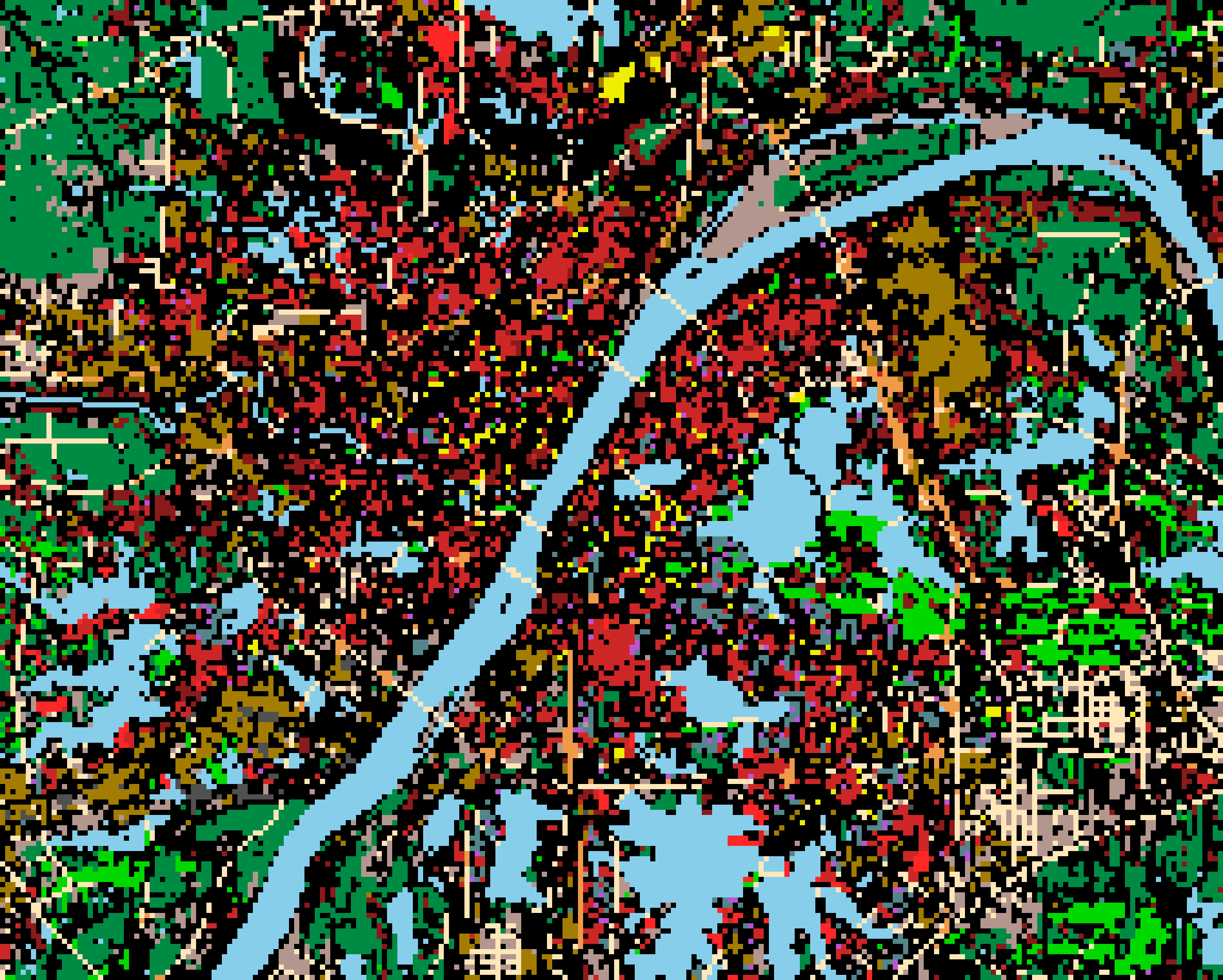}
		\label{label_2014}
	}
	\hspace{-0.15in}
	\subfigure[] {
		\includegraphics[width=85mm]{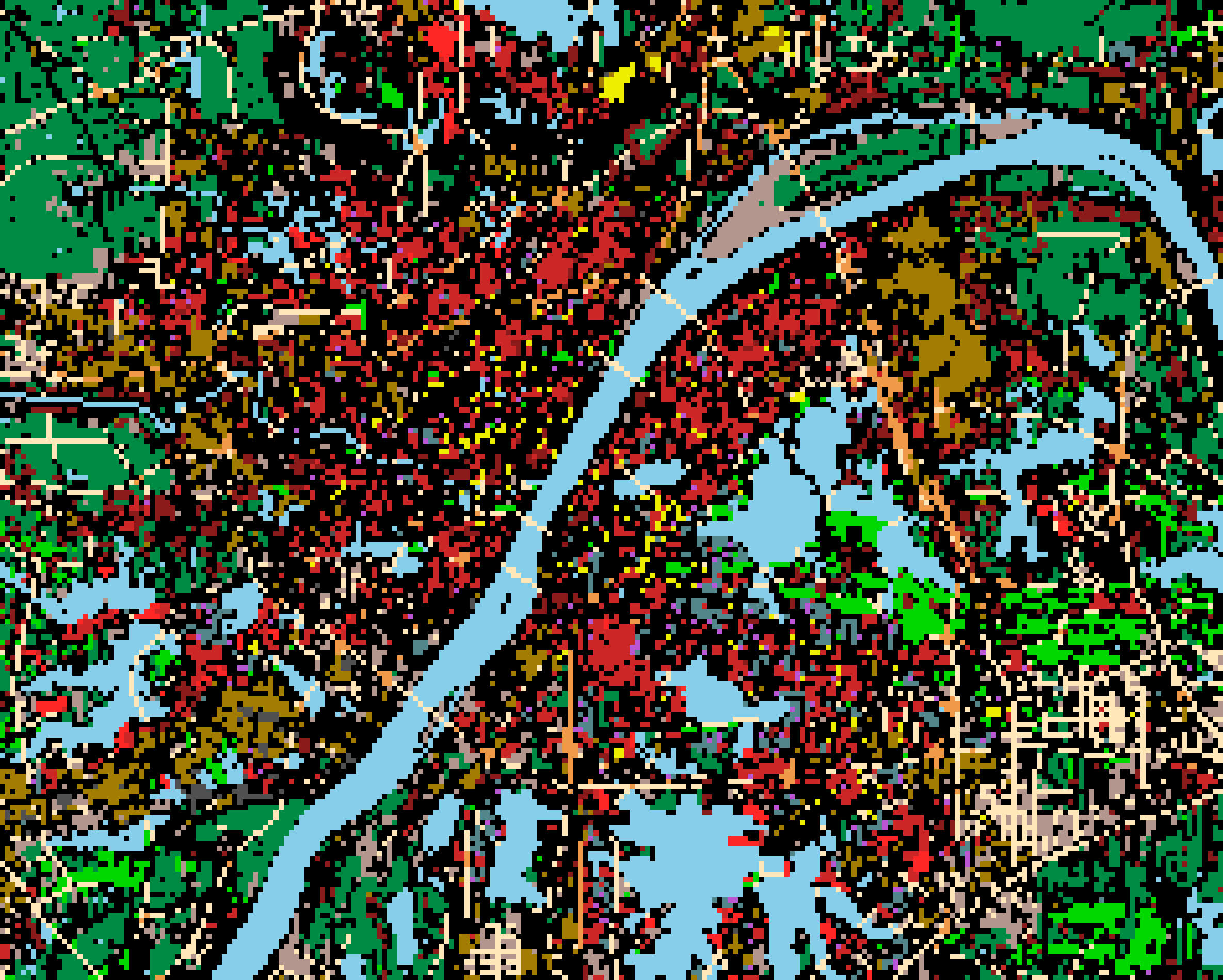}
		\label{label_2016}
	}
	\hspace{-0.15in}

	\vspace{-0.14in}
	\subfigure{
		\hspace{0.08in}
		\includegraphics[scale=0.52]{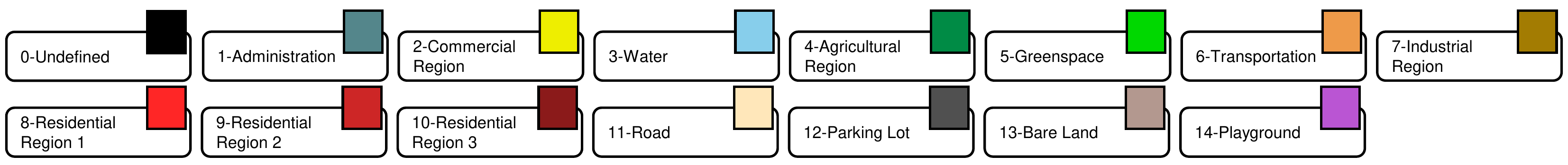}
	}
	\caption{The large size high-resolution images of Wuhan dataset. They are acquired in (a)2014 and (b)2016, respectively. Image (c) and (d) respectively present the spatial distribution of the scene categories. The undefined patches are colored in black.}
	\label{fig_wuhan}
\end{figure*}
\begin{figure*}[h!t]
	\subfigure[] {
		\includegraphics[scale=0.24]{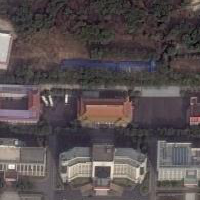}
		\includegraphics[scale=0.24]{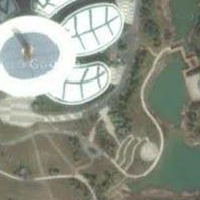}
	}
	\hspace{-0.15in}
	\subfigure[] {
		\includegraphics[scale=0.24]{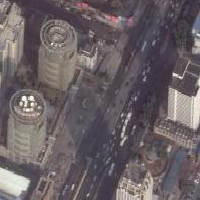}
		\includegraphics[scale=0.24]{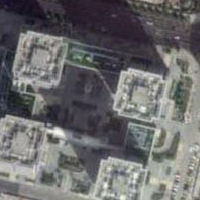}
	}
	\hspace{-0.15in}
	\subfigure[] {
		\includegraphics[scale=0.24]{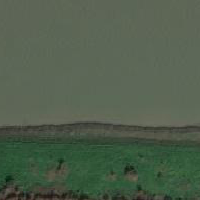}
		\includegraphics[scale=0.24]{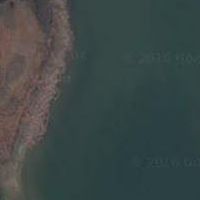}
	}
	\hspace{-0.15in}
	\subfigure[] {
		\includegraphics[scale=0.24]{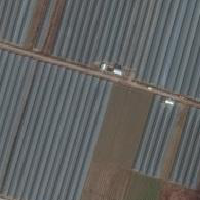}
		\includegraphics[scale=0.24]{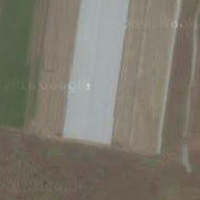}
	}
	\hspace{-0.15in}
	\subfigure[] {
		\includegraphics[scale=0.24]{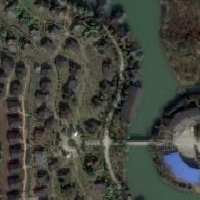}
		\includegraphics[scale=0.24]{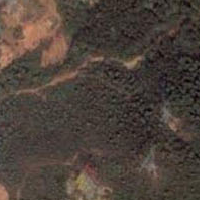}
	}
	\hspace{-0.15in}

	\subfigure[] {
		\includegraphics[scale=0.24]{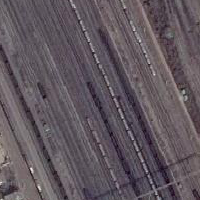}
		\includegraphics[scale=0.24]{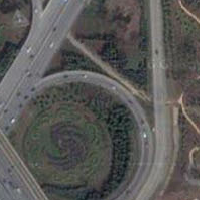}
	}
	\hspace{-0.15in}
	\subfigure[] {
		\includegraphics[scale=0.24]{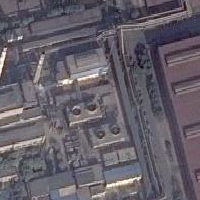}
		\includegraphics[scale=0.24]{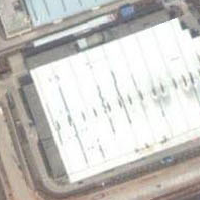}
	}
	\hspace{-0.15in}
	\subfigure[] {
		\includegraphics[scale=0.24]{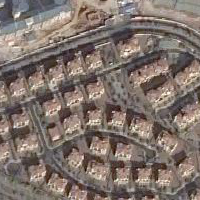}
		\includegraphics[scale=0.24]{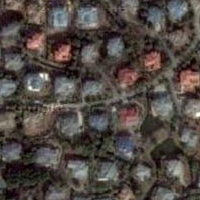}
	}
	\hspace{-0.15in}
	\subfigure[] {
		\includegraphics[scale=0.24]{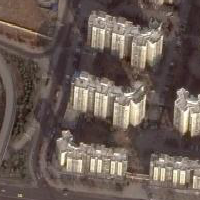}
		\includegraphics[scale=0.24]{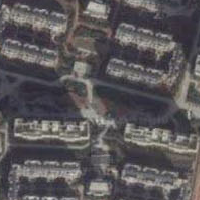}
	}
	\hspace{-0.15in}
	\subfigure[] {
		\includegraphics[scale=0.24]{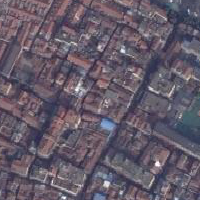}
		\includegraphics[scale=0.24]{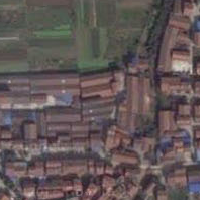}
	}
	\hspace{-0.15in}

	\subfigure[] {
		\includegraphics[scale=0.24]{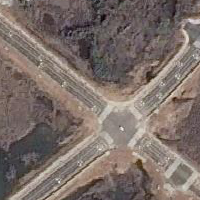}
		\includegraphics[scale=0.24]{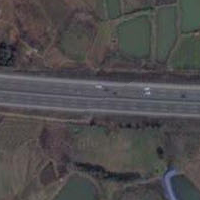}
	}
	\hspace{-0.15in}
	\subfigure[] {
		\includegraphics[scale=0.24]{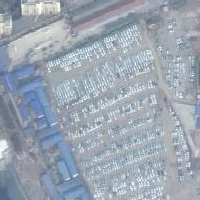}
		\includegraphics[scale=0.24]{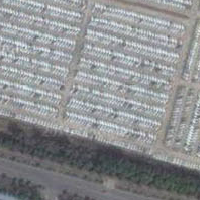}
	}
	\hspace{-0.15in}
	\subfigure[] {
		\includegraphics[scale=0.24]{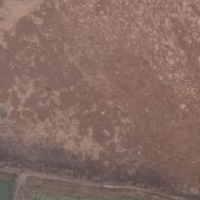}
		\includegraphics[scale=0.24]{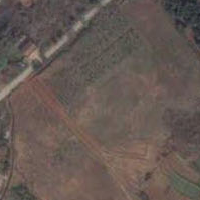}
	}
	\hspace{-0.15in}
	\subfigure[] {
		\includegraphics[scale=0.24]{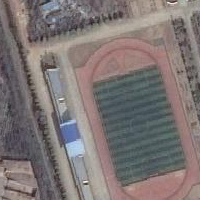}
		\includegraphics[scale=0.24]{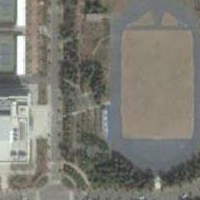}
	}
	\caption{Typical samples of (a) Administration, (b) Commercial Region, (c) Water, (d) Farmland, (e) Greenspace, (f) Transportation, (g) Industrial Region, (h) Residential Region 1, (i) Residential Region 2, (j) Residential Region 3, (k) Road, (l)Parking Lot, (m) Bare Land and (n) Playground.}
	\label{fig_samples}
\end{figure*}

\par As shown in Fig~\ref{wuhan_2014} and \ref{wuhan_2016}, Wuhan dataset contains two large size high-resolution images with spatial resolution of 2m covering Wuhan city, Hubei Province, China. The images are respectively obtained in 2014 and 2016 and have a spatial size of $47537\times 38100$. We divided each large size image into $200\times 200$ non-overlapping image patches, and assigned each patch with a specific land-use scene category label by visual interpretation. Except the images that were too complex to recognize the scene categories, denoted by 0-Undefined in Fig~\ref{fig_wuhan}, we obtained 23555 labeled images with 14 scene categories for each time. The spatial distribution of all the labeled images are presented in Fig~\ref{label_2014} and \ref{label_2016}. Some typical samples for each scene category are presented in Fig~(\ref{fig_samples}).

\begin{table*}[]
	\caption{The number of each scene category and changes in 2014 and 2016 of Wuhan Dataset.}
	\renewcommand{\arraystretch}{1.2}
	\label{wuhan_2014_2016}
	\centering
	\begin{tabular}{|c|c|cccccccccccccc|c|}

		\hline
		\multicolumn{2}{|c|}{\multirow{2}{*}{}} & \multicolumn{14}{c|}{\textbf{2016}} & \multirow{2}{*}{\textbf{sum}}                                                                                                                                                                                 \\ \cline{3-16}
		\multicolumn{2}{|c|}{}                  & \textbf{1}                          & \textbf{2}                    & \textbf{3} & \textbf{4} & \textbf{5} & \textbf{6} & \textbf{7} & \textbf{8} & \textbf{9} & \textbf{10} & \textbf{11} & \textbf{12} & \textbf{13} & \textbf{14} &              \\ \hline
		\multirow{14}{*}{\textbf{2014}}         & \textbf{1}                          & 465                           & 0          & 0          & 0          & 0          & 0          & 0          & 0          & 0           & 0           & 0           & 0           & 0           & 0     & 465  \\
		                                        & \textbf{2}                          & 0                             & 268        & 0          & 0          & 0          & 0          & 0          & 0          & 0           & 0           & 0           & 0           & 0           & 0     & 268  \\
		                                        & \textbf{3}                          & 0                             & 0          & 5759       & 3          & 0          & 0          & 0          & 0          & 0           & 0           & 15          & 0           & 3           & 0     & 5780 \\
		                                        & \textbf{4}                          & 0                             & 0          & 1          & 4528       & 15         & 0          & 0          & 0          & 0           & 0           & 28          & 0           & 21          & 0     & 4593 \\
		                                        & \textbf{5}                          & 0                             & 0          & 0          & 2          & 1083       & 0          & 0          & 0          & 0           & 0           & 1           & 0           & 10          & 0     & 1096 \\
		                                        & \textbf{6}                          & 0                             & 1          & 0          & 0          & 0          & 409        & 0          & 0          & 0           & 0           & 0           & 0           & 0           & 0     & 410  \\
		                                        & \textbf{7}                          & 0                             & 0          & 0          & 0          & 0          & 0          & 1965       & 0          & 1           & 0           & 0           & 2           & 4           & 0     & 1972 \\
		                                        & \textbf{8}                          & 0                             & 0          & 0          & 0          & 0          & 0          & 0          & 264        & 0           & 0           & 0           & 0           & 0           & 0     & 264  \\
		                                        & \textbf{9}                          & 1                             & 0          & 0          & 0          & 0          & 0          & 0          & 0          & 2642        & 0           & 0           & 0           & 3           & 1     & 2647 \\
		                                        & \textbf{10}                         & 0                             & 0          & 0          & 0          & 0          & 0          & 0          & 0          & 1           & 1345        & 2           & 0           & 12          & 0     & 1360 \\
		                                        & \textbf{11}                         & 1                             & 1          & 0          & 0          & 0          & 3          & 7          & 0          & 4           & 0           & 2551        & 1           & 1           & 0     & 2569 \\
		                                        & \textbf{12}                         & 0                             & 0          & 0          & 0          & 0          & 0          & 2          & 0          & 0           & 0           & 1           & 130         & 0           & 0     & 133  \\
		                                        & \textbf{13}                         & 1                             & 6          & 34         & 69         & 10         & 0          & 20         & 0          & 33          & 0           & 88          & 6           & 1583        & 1     & 1851 \\
		                                        & \textbf{14}                         & 0                             & 0          & 0          & 0          & 0          & 0          & 0          & 0          & 0           & 0           & 0           & 0           & 0           & 147   & 147  \\ \hline
		\multicolumn{2}{|c|}{\textbf{sum}}      & 468                                 & 276                           & 5794       & 4602       & 1108       & 412        & 1994       & 264        & 2681       & 1345        & 2686        & 139         & 1637        & 149         & 23555        \\ \hline
	\end{tabular}

\end{table*}

\par In Tab~\ref{wuhan_2014_2016}, we showed the detailed statistics of the number of each category in 2014 and 2016 and the number of their changes. It's noted that Water and Farmland include about 10000 images in total while each of Parking Lot and Playground only contains about 150 images. Therefore, due to the imbalance among categories, this dataset is quite challenging. In the experiments, we randomly split 70\%, 10\% and 20\% respectively as training, validation and testing set.

\subsection{Experiment Settings}
\par In the training procedure, we used a Momentum Optimizer \cite{sutskever2013importance} with initial learning rate of 0.001 and momentum of 0.9. The number of epochs and batchsize in training are respectively set as 100 and 32. We also used an $\mathcal{L}_2$ regularization term with weights of 0.0001 in training. The convolution module of CorrFusionNet could be any classification network. In this work, we tried several mostly used architectures for image classification, including VGGNet \cite{Simonyan2014}, InceptionV3 \cite{Szegedy2015a}, ResNet\cite{He2016} and DenseNet \cite{Huang2017}. For the fully connected layers of CorrFusionNet, we set two layers with 1024 units for each view. The activation functions of fully connected layers are all set as $ReLU$ \cite{glorot2011deep}. The Xavier initializer \cite{glorot2010understanding} is utilized to initialize the weights in all layers of CorrFusionNet. The hyperparameters involved include dimensionality-reduction ratio $r$ in Eq.(\ref{x_fc}) and the momentum parameter $\rho$ in Eq.(\ref{stable_cov}). They are discussed and presented on the validation set in Section~\ref{hyperparameter}. Our implementation code is available on Github.\footnote{\href{https://github.com/rulixiang/CorrFusionNet}{https://github.com/rulixiang/CorrFusionNet}}

\subsection{Evaluation Criteria}
\par We mainly used overall accuracy as the evaluation criterion in experiments. Assuming the predicted and true label of time 1 and time 2 are $P_{t1}, P_{t2}, L_{t1}, L_{t2} \in \mathbb{R}^n$, respectively, where $n$ is the number of samples. OA\_t1, OA\_t2, OA\_bi and OA\_tr are defined as
\begin{equation}
	\label{eval}
	\begin{split}
		OA\_t1&= \frac{1}{n}\sum_{i=1}^n(P_{t1}(i)==L_{t1}(i)) \\
		OA\_t2&= \frac{1}{n}\sum_{i=1}^n(P_{t2}(i)==L_{t2}(i))\\
		OA\_bi&= \frac{1}{n}\sum_{i=1}^n((P_{t1}(i)==P_{t2}(i))==(L_{t1}(i)==L_{t2}(i)))\\
		OA\_tr&= \frac{1}{n}\sum_{i=1}^n((P_{t1}(i)==L_{t1}(i)) \& (P_{t2}(i)==L_{t2}(i)))
	\end{split}
\end{equation}
As shown in Eq.(\ref{eval}), the evaluation criteria include overall accuracies of the scene classification of time 1 and time 2, the binary change detection (change/unchange) and the transition change detection (from-to), which are correspondingly denoted by OA\_t1, OA\_t2, OA\_bi and OA\_tr.
\subsection{Experimental Results}
\par To find the most suitable backbone network for Wuhan dataset, we performed experiments and evaluated accuracies on the validation set using several common image classification networks. The results are presented in Tab~\ref{tab_wuhan_val_acc}. The best and second-best values of each column are respectively highlighted in bold and underlined. 

\begin{table}[]
	\renewcommand{\arraystretch}{1.2}
	\centering

	\caption{Scene classification and change detection accuracies on the validation set of Wuhan dataset.}
	\begin{tabular}{l|cccc}
		\hline
		                     & \textbf{OA\_t1}     & \textbf{OA\_t2}     & \textbf{OA\_bi}     & \textbf{OA\_tr}     \\ \hline
		\textbf{VGG16}       & 89.17\%             & 88.58\%             & 85.31\%             & 82.38\%             \\
		\textbf{VGG19}       & 86.79\%             & 86.92\%             & 84.29\%             & 79.92\%             \\ \hline
		\textbf{InceptionV3} & 87.01\%             & 88.58\%             & 84.97\%             & 80.85\%             \\ \hline
		\textbf{ResNet50}    & 85.19\%             & 86.76\%             & 83.02\%             & 78.48\%             \\
		\textbf{ResNet101}   & 83.40\%             & 86.67\%             & 81.10\%             & 76.65\%             \\
		\textbf{ResNet152}   & 82.00\%             & 85.01\%             & 79.07\%             & 74.95\%             \\ \hline
		\textbf{DenseNet121} & \underline{90.06\%} & \textbf{90.23}\%    & \textbf{88.11}\%    & \textbf{84.59}\%    \\
		\textbf{DenseNet169} & \textbf{90.40\%}    & 89.94\%             & \underline{87.81\%} & \underline{84.33\%} \\
		\textbf{DenseNet201} & 89.81\%             & \underline{90.19\%} & 87.47\%             & 84.12\%             \\ \hline
	\end{tabular}
	\label{tab_wuhan_val_acc}
\end{table}
\par As could be observed in Tab~\ref{tab_wuhan_val_acc}, DenseNet achieved the highest accuracies, followed by VGGNet. Specifically, DenseNet121 and DenseNet169 had the best and second best performance in general, respectively. Furthermore, for a specific architecture, deeper models didn't bring higher accuracies. We think it's because of the over-parameterization of deeper models. Based on the results in Tab~\ref{tab_wuhan_val_acc}, to verify the generality of our CorrFusion module, we chose VGG16, InceptionV3, ResNet50 and DenseNet121 as backbone networks for the following experiments.
\begin{table}[]
	\renewcommand{\arraystretch}{1.2}
	\centering
	\caption{Scene classification accuracies on the testing set of Wuhan dataset and Hanyang Dataset.}
	\begin{tabular}{l|c|c|c|c}
		\hline
		\textbf{}                         & \multicolumn{2}{c|}{\textbf{Wuhan}}  & \multicolumn{2}{c}{\textbf{Hanyang}}                                                                              \\ \cline{2-5}
		\textbf{}                         & \multicolumn{1}{c|}{\textbf{OA\_t1}} & \multicolumn{1}{c|}{\textbf{OA\_t2}} & \multicolumn{1}{c|}{\textbf{OA\_t1}} & \multicolumn{1}{c}{\textbf{OA\_t2}} \\ \hline
		\textbf{BoVW\cite{Wu2016}}        & 80.30\%                              & 85.46\%                              & 80.29\%                              & 80.19\%                             \\
		\textbf{BoVW + KSFA\cite{Wu2017}} & 81.72\%                              & 84.00\%                              & 85.52\%                              & 88.95\%                             \\
		\textbf{DCCANet\cite{Wang2019}}   & 86.36\%                              & 88.11\%                              & 84.50\%                              & 88.20\%                             \\ \hline
		\textbf{VGG16}                    & 89.35\%                              & 89.43\%                              & 85.07\%                              & 84.56\%                             \\
		\textbf{VGG16 + CorrFusion}       & \textbf{91.03\%}                     & \textbf{92.21\%}                     & \underline{86.29\%}                  & 86.38\%                             \\ \hline
		\textbf{InceptionV3}              & 87.63\%                              & 88.88\%                              & 78.95\%                              & 84.48\%                             \\
		\textbf{InceptionV3 + CorrFusion} & 88.90\%                              & 88.71\%                              & 79.71\%                              & 85.43\%                             \\ \hline
		\textbf{ResNet50}                 & 86.08\%                              & 88.13\%                              & 81.64\%                              & 87.62\%                             \\
		\textbf{ResNet50 + CorrFusion}    & 88.57\%                              & 89.13\%                              & 81.86\%                              & 89.05\%                             \\ \hline
		\textbf{DenseNet121}              & 89.75\%                              & 90.28\%                              & 85.62\%                              & \underline{89.14\%}                 \\
		\textbf{DenseNet121 + CorrFusion} & \underline{90.85\%}                  & \underline{92.08\%}                  & \textbf{86.42\%}                     & \textbf{89.70\%}                    \\ \hline
	\end{tabular}
	\label{tab_acc_cls}
\end{table}
\par The multi-temporal scene classification results of Wuhan and Hanyang dataset are presented in Tab~\ref{tab_acc_cls}. We compared the accuracies of VGG16, InceptionV3, ResNet50 and DenseNet121 with and without proposed CorrFusion module. Some previous methods are also included. The presented results of proposed CorrFusionNet are obtained with $r$ and $\rho$ respectively fixed to 2 and 0.9. On Wuhan dataset, VGG16 with CorrFusion module, namely CorrFusionNet with VGG16 as backbone network, achieved the highest scene classification accuracies on both time 1 and time 2. DenseNet121 with CorrFusion also obtained very high performance. Both VGG16 and DenseNet121 largely overpassed existing scene change detection methods. It's also noticed that our proposed CorrFusion remarkably improved the classification performance of all backbone networks. Similarly, on Hanyang dataset, our method could also improve the scene classification accuracies of tested backbone networks and surpass DCCANet and BoVW based methods.

\begin{figure*}[h!t] \centering
	\subfigure[] {
		\label{fig_confusionmat_2014}
		\includegraphics[width=85mm]{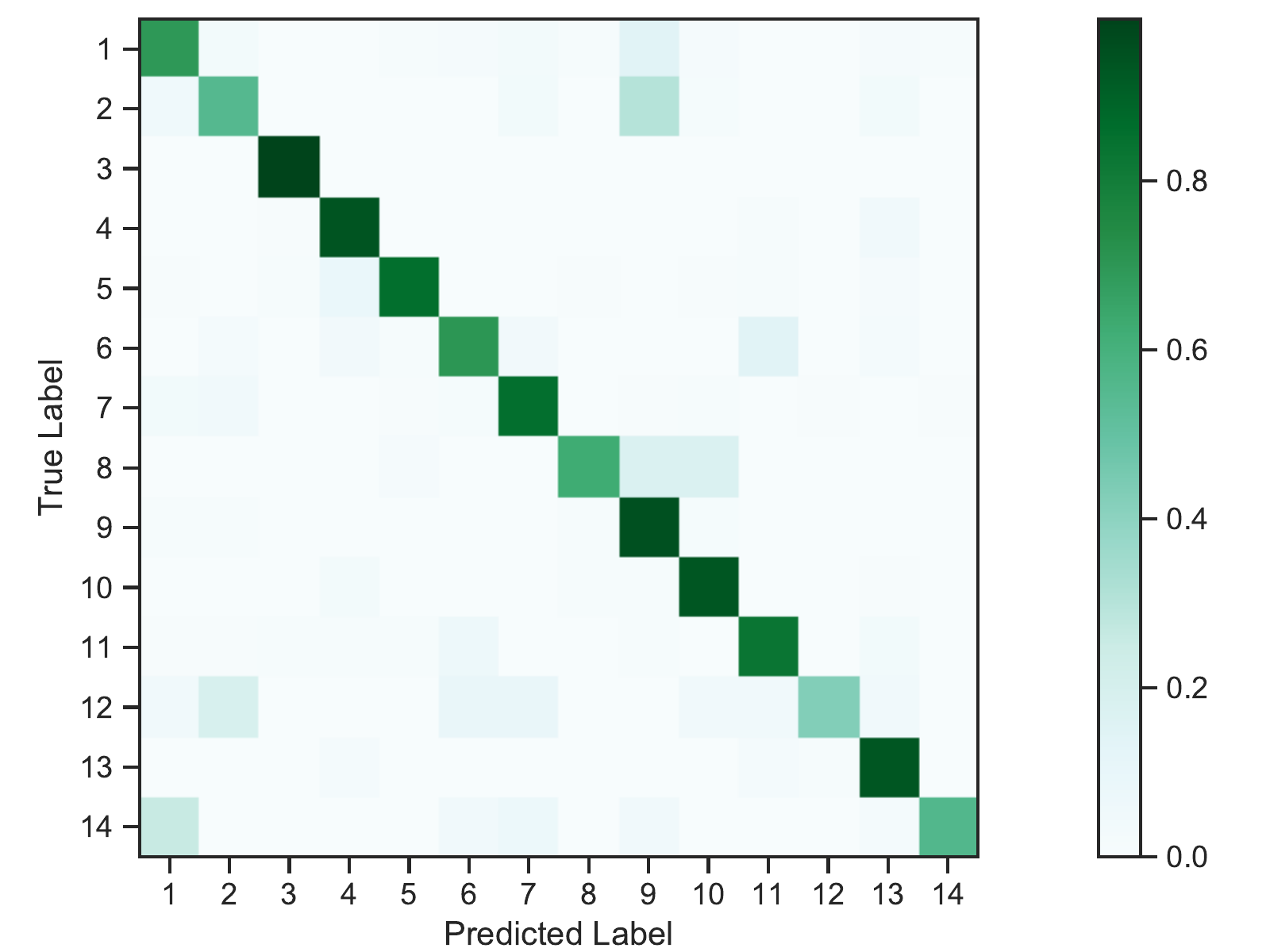}
	}
	\subfigure[] {
		\label{fig_confusionmat_2016}
		\includegraphics[width=85mm]{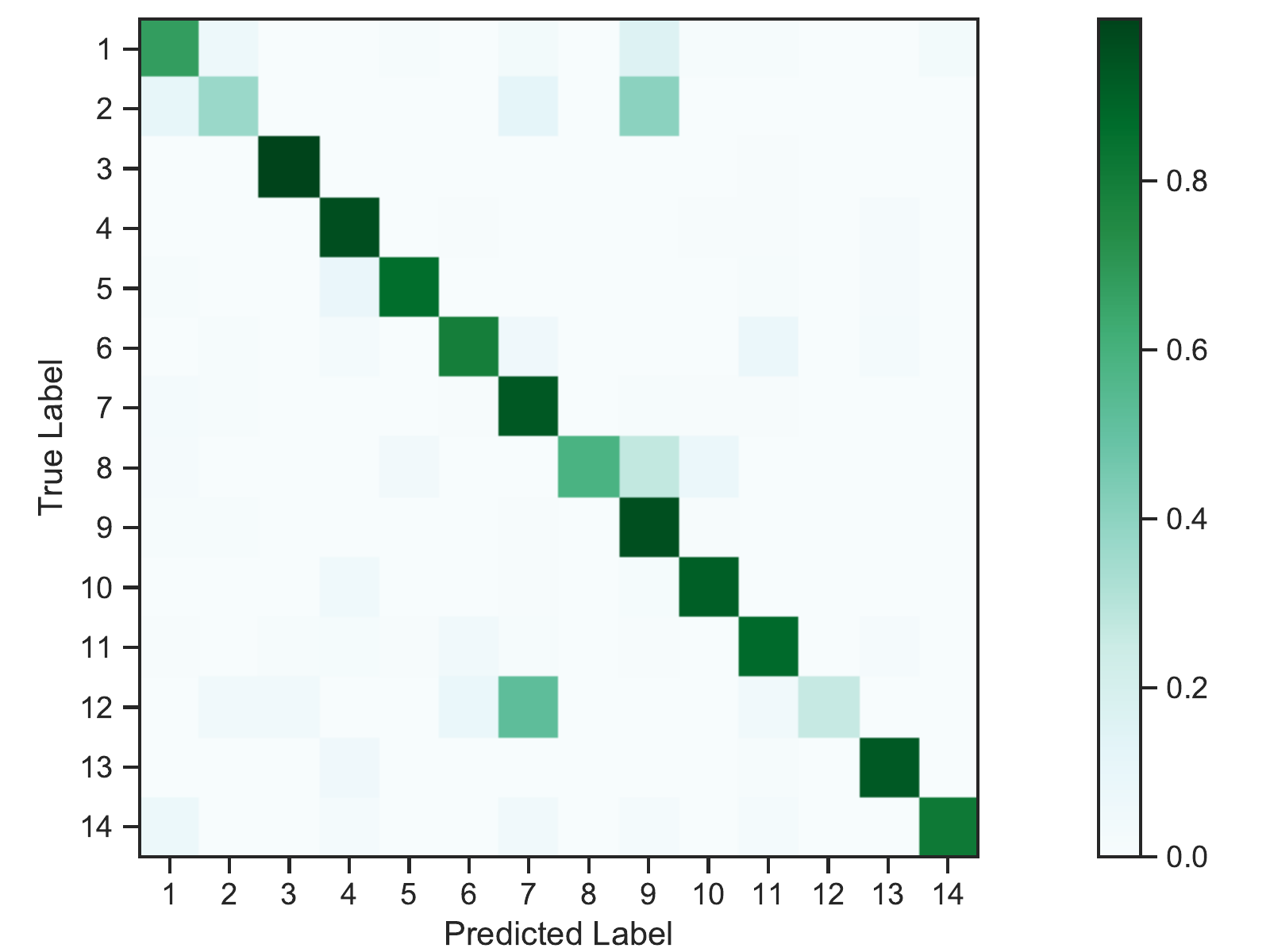}
	}

	\caption{The confusion matrices of scene classification results on the testing set, with (a) is the result in 2014 and (b) is the result in 2016. Deeper color indicates higher classification accuracy.}
	\label{fig_confusionmat}
\end{figure*}
\par In Fig~\ref{fig_confusionmat}, we presented the confusion matrices of scene classification results using VGG16 with CorrFusion on the testing set. Each row in Fig~\ref{fig_confusionmat_2014} and Fig~\ref{fig_confusionmat_2016} respectively denotes the predicted results of the samples from a specific scene category in 2014 and 2016. As shown, classes with larger number of images could achieve much higher classification accuracies, while categories with few images only achieved much worse accuracies, such as 2-Commercial and 12-Parking. Besides, the inter-class correlation also resulted in the low classification accuracies of specific classes. For example, the similarity between 2-Commercial and 9-Residential-2 is one of the reasons for the poor performance on classifying 2-Commercial.

\begin{figure*}[h!t] \centering
	\subfigure[] {
		\includegraphics[width=85mm]{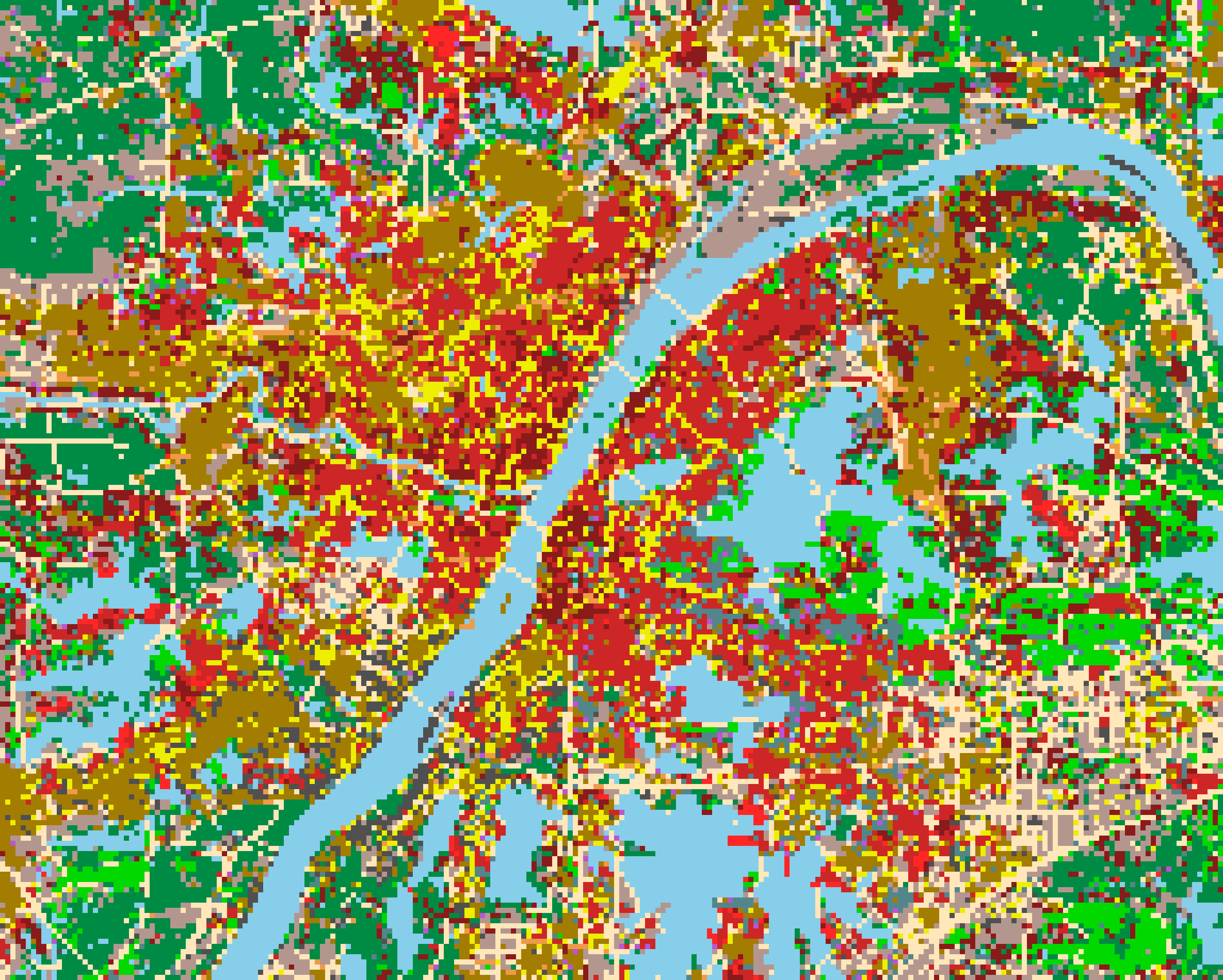}
	}
	\hspace{-0.15in}
	\subfigure[] {
		\includegraphics[width=85mm]{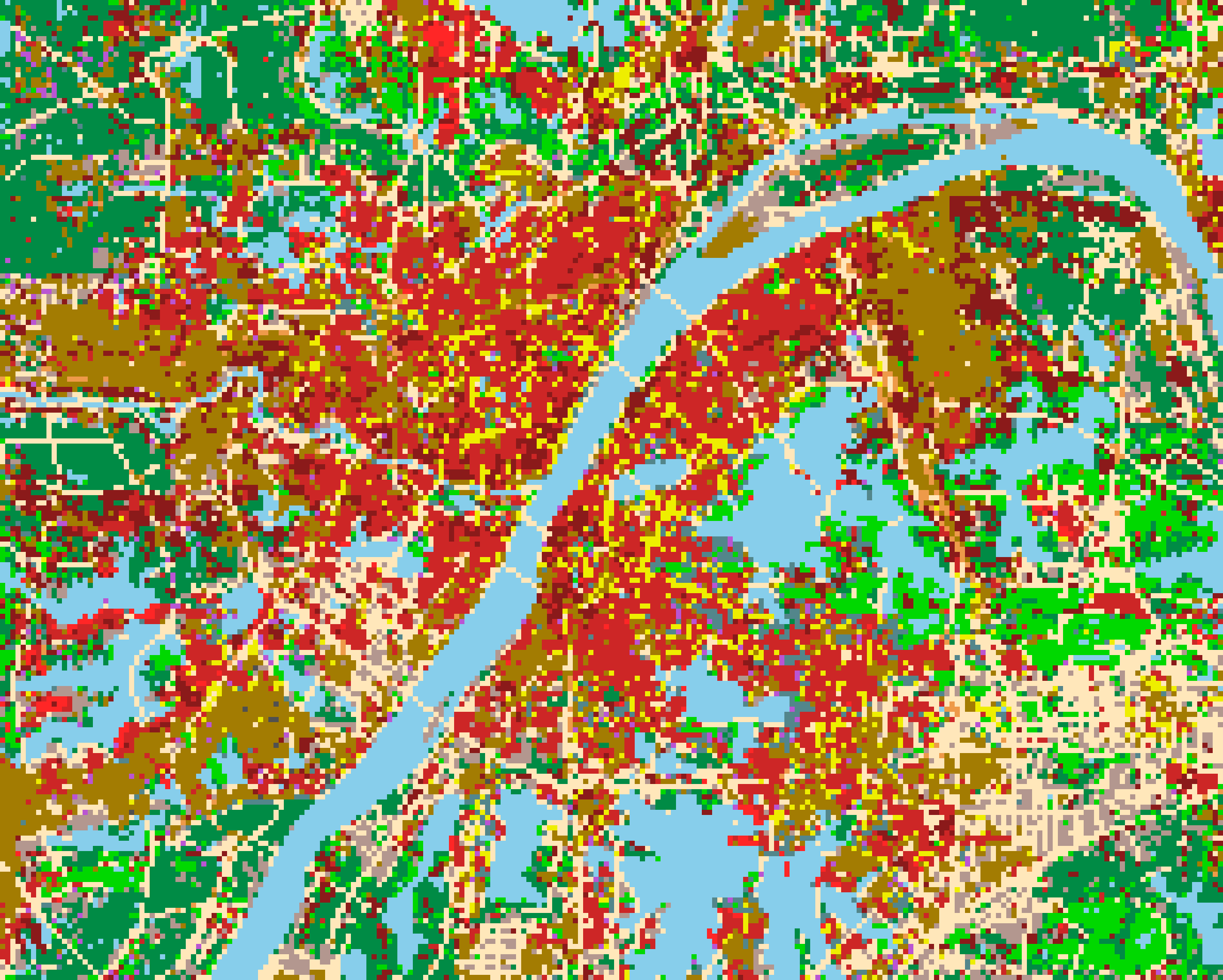}
	}
	\hspace{-0.15in}

	\vspace{-0.14in}
	\subfigure{
		\hspace{0.03in}
		\includegraphics[scale=0.56]{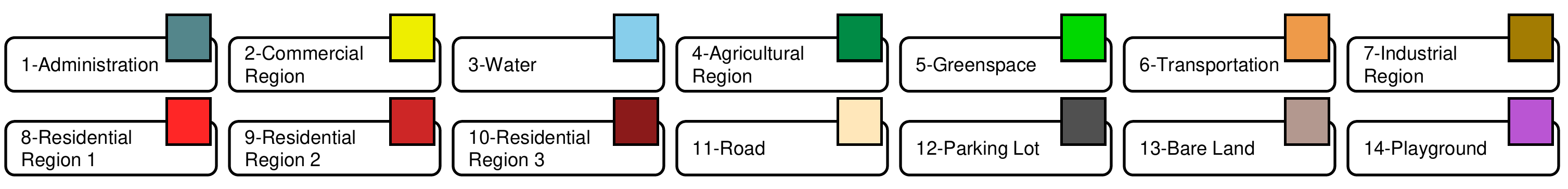}
	}

	\caption{The predicted maps on our dataset using a trained CorrFusionNet.}
	\label{fig_prediction}
\end{figure*}
\par We predicted all the image patches in Fig~\ref{fig_wuhan} and presented the final results in Fig~\ref{fig_prediction}. Most of the predictions agreed with the annotations in Fig~\ref{label_2014} and Fig~\ref{label_2016}. However, there're still some apparent mistakes, such as classifying Bare Land as Industrial Region.

\begin{table}[]
	\renewcommand{\arraystretch}{1.2}
	\centering
	\caption{Scene change detection accuracies on the testing set of Wuhan dataset and Hanyang Dataset.}
	\begin{tabular}{l|c|c|c|c}
		\hline
		\textbf{}                         & \multicolumn{2}{c|}{\textbf{Wuhan}}  & \multicolumn{2}{c}{\textbf{Hanyang}}                                                                              \\ \cline{2-5}
		\textbf{}                         & \multicolumn{1}{c|}{\textbf{OA\_bi}} & \multicolumn{1}{c|}{\textbf{OA\_tr}} & \multicolumn{1}{c|}{\textbf{OA\_bi}} & \multicolumn{1}{c}{\textbf{OA\_tr}} \\ \hline
		\textbf{BoVW \cite{Wu2016}}       & 79.03\%                              & 73.90\%                              & {79.01\%}                            & 66.00\%                             \\
		\textbf{BoVW + KSFA\cite{Wu2017}} & 89.76\%                              & 77.88\%                              & {84.76\%}                            & 77.24\%                             \\
		\textbf{DCCANet\cite{Wang2019}}   & 84.32\%                              & 80.34\%                              & \underline{87.04\%}                  & 76.80\%                             \\ \hline
		\textbf{VGG16}                    & 86.12\%                              & 83.21\%                              & {83.38\%}                            & 73.19\%                             \\
		\textbf{VGG16 + CorrFusion}       & 92.34\%                              & \underline{88.24\%}                  & {85.33\%}                            & 75.71\%                             \\ \hline
		\textbf{InceptionV3}              & 84.02\%                              & 81.13\%                              & {79.52\%}                            & 67.14\%                             \\
		\textbf{InceptionV3 + CorrFusion} & \textbf{93.29\%}                     & 86.40\%                              & {81.24\%}                            & 70.48\%                             \\ \hline
		\textbf{ResNet50}                 & 83.92\%                              & 80.02\%                              & {83.15\%}                            & 72.72\%                             \\
		\textbf{ResNet50 + CorrFusion}    & 90.89\%                              & 85.24\%                              & {85.30\%}                            & 74.76\%                             \\ \hline
		\textbf{DenseNet121}              & 88.39\%                              & 84.72\%                              & {86.19\%}                            & \underline{77.33\%}                 \\
		\textbf{DenseNet121 + CorrFusion} & \underline{92.57\%}                  & \textbf{88.29\%}                     & \textbf{88.13\%}                     & \textbf{78.85\%}                    \\ \hline
	\end{tabular}
	\label{tab_cd}
\end{table}
\par We obtained the scene change detection results of Wuhan and Hanyang dataset by post-classification comparison. As presented in Tab~\ref{tab_cd}, the proposed CorrFusion module significantly improved the binary and transition scene change detection accuracies. Particularly, on Wuhan dataset, our method brought an improvement by $\sim 5\%$ for both OA\_bi and OA\_tr. We think the reason is Wuhan dataset contains much more unchanged scene pairs so that the learned weights in CorrFusion module are more reliable. In contrast, the training set of Hanyang dataset is small, so the CorrFusion module cannot be effectively learned, which leads to the accuracy improvements on Hanyang dataset are not as remarkable as on Wuhan dataset.
\begin{table}[]
	\renewcommand{\arraystretch}{1.2}
	\caption{The True Positive, True Negative, False Positive and False Negative samples on the testing set of Wuhan dataset with different methods.}
	\label{tab_tp}
	\centering
	\begin{tabular}{l|c|c|c|c}
		\hline
		                            & \textbf{TP} & \textbf{FN} & \textbf{FP}  & \textbf{TN}   \\ \hline
		\textbf{VGG16}              & \textbf{69} & \textbf{13} & 641          & 3989          \\ \hline
		\textbf{VGG16 + DCCA}       & {68}        & {14}        & 649          & 3981          \\ \hline
		\textbf{VGG16 + Soft DCCA}  & 67          & 15          & 586          & 4044          \\ \hline
		\textbf{VGG16 + CorrFusion} & 67          & 15          & \textbf{355} & \textbf{4275} \\ \hline
	\end{tabular}
\end{table}
\par To verify the effect of the proposed module, we compared the numbers of True Positive, True Negative, False Positive and False Negative samples on the testing set with VGG16 and VGG16 with different modules. The results are presented in Tab~\ref{tab_tp}. It could be observed that VGG16 with DCCA module achieved almost the same performance with VGG16, which indicated that DCCA could barely work for large-scale dataset. In contrast, our proposed CorrFusionNet performs much better on FP and TN but is slightly worse on TP and FN. The reason is that the distances between changed scene pairs are not explicitly constrained in the objective function. Due to the very imbalance between the changed and unchanged scene pairs, the learned weights in Eq.(\ref{xy_w}) for changed and unchanged pairs will be all close to 1. However, a larger $\mathbf{w}$ could help to improve the representation ability of unchanged scene images (FP, TN), but will instead weaken the performance on classifying changed image pairs (TP, FN) because of the fusion of bi-temporal features from different categories.
\subsection{Hyperparameter Analysis}
\label{hyperparameter}
\par The results presented above are all obtained with dimensionality-reduction ratio $r$ of 2 and momentum parameter $\rho$ of 0.9. In this section, we will show how they'll impact on the final results of multiple-temporal scene classification and scene change detection.
\subsubsection{Dimensionality-reduction ratio}
\label{ratio}
\par We firstly showed the classification and change detection accuracies on the validation set by respectively setting the dimensionality-reduction ratio $r$ to $[1, 2, 4, 8, 16]$.
\begin{figure}[h!t] \centering
	\includegraphics[scale=0.33]{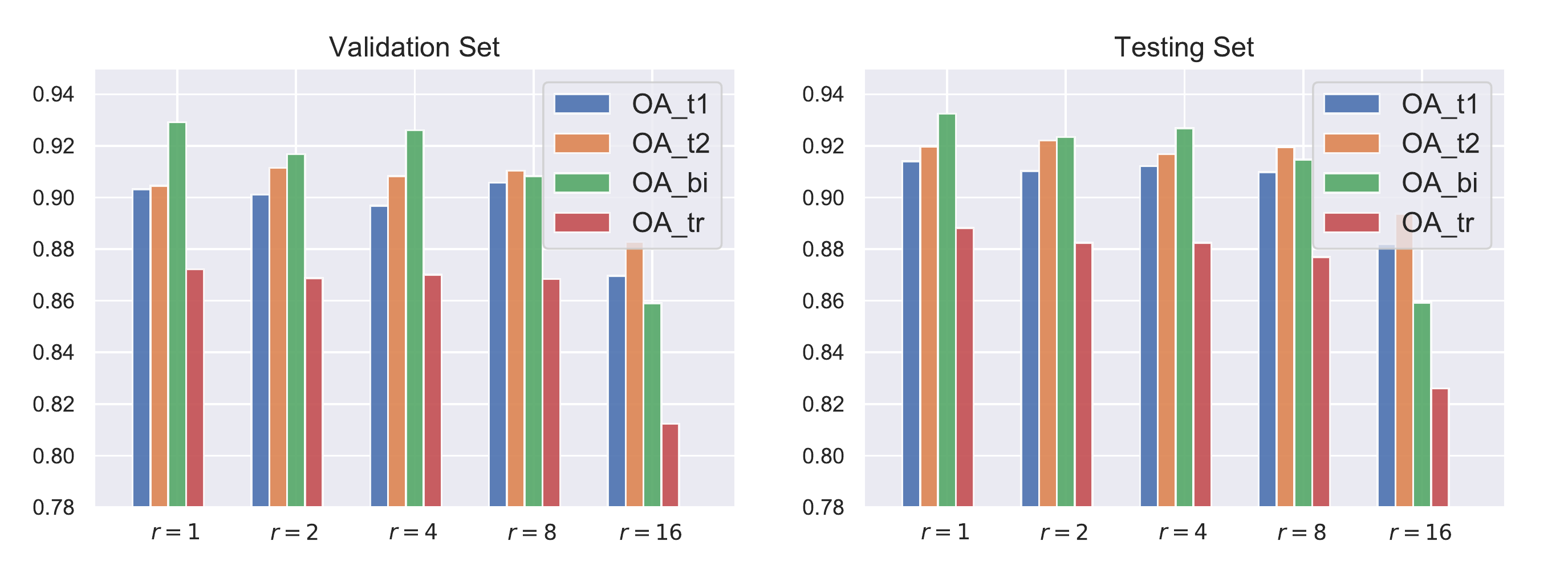}
	\caption{The scene classification and scene change detection accuracies with $r=1$, $r=2$, $r=4$, $r=8$ and $r=16$.}
	\label{fig_ratio}
\end{figure}
\par As presented in Fig~(\ref{fig_ratio}), CorrFusionNet with $r=2,4$ and 8 could achieve comparative performances with $r=1$ with fewer parameters, while CorrFusionNet with $r=16$ obtained the lowest accuracies on all 4 evaluation criteria. We account that the proposed CorrFusion module with the dimensionality-reduction ratio of 2, 4 and 8 could still retain the principal and correlated components of the bi-temporal feature embeddings. But a lower ratio will probably result in the reduction of the information and thus decrease the accuracies.
\begin{table}[]
	\centering
	\renewcommand{\arraystretch}{1.2}
	\caption{The comparison of number of parameters and accuracies of CorrFusion module with different $r$.}
	\label{tab_para}
	\begin{tabular}{l|ccc}
		\hline
		                           & \textbf{Parameters} & \textbf{validation} & \textbf{testing} \\ \hline
		\textbf{CorrFusion $r=1$}  & 4.01 M              & 87.22\%             & 88.82\%          \\
		\textbf{CorrFusion $r=2$}  & 2.00 M              & 86.88\%             & 88.24\%          \\
		\textbf{CorrFusion $r=4$}  & 1.00 M              & 87.01\%             & 88.24\%          \\
		\textbf{CorrFusion $r=8$}  & 0.50 M              & 86.84\%             & 87.69\%          \\
		\textbf{CorrFusion $r=16$} & 0.25 M              & 81.23\%             & 82.60\%          \\ \hline
	\end{tabular}
\end{table}
\par In Tab~\ref{tab_para}, we compared the number of parameters and performance of CorrFusion module with different $r$ on Wuhan dataset. It could be clearly observed that CorrFusionNet with $r=1$ only slightly outperform other methods with much more parameters. CorrFusionNet with $r=2,4$ and 8 achieved fairly accuracies with fewer parameters.
\subsubsection{Momentum parameter $\rho$}
\begin{figure}[h!t] \centering
	\includegraphics[scale=0.32]{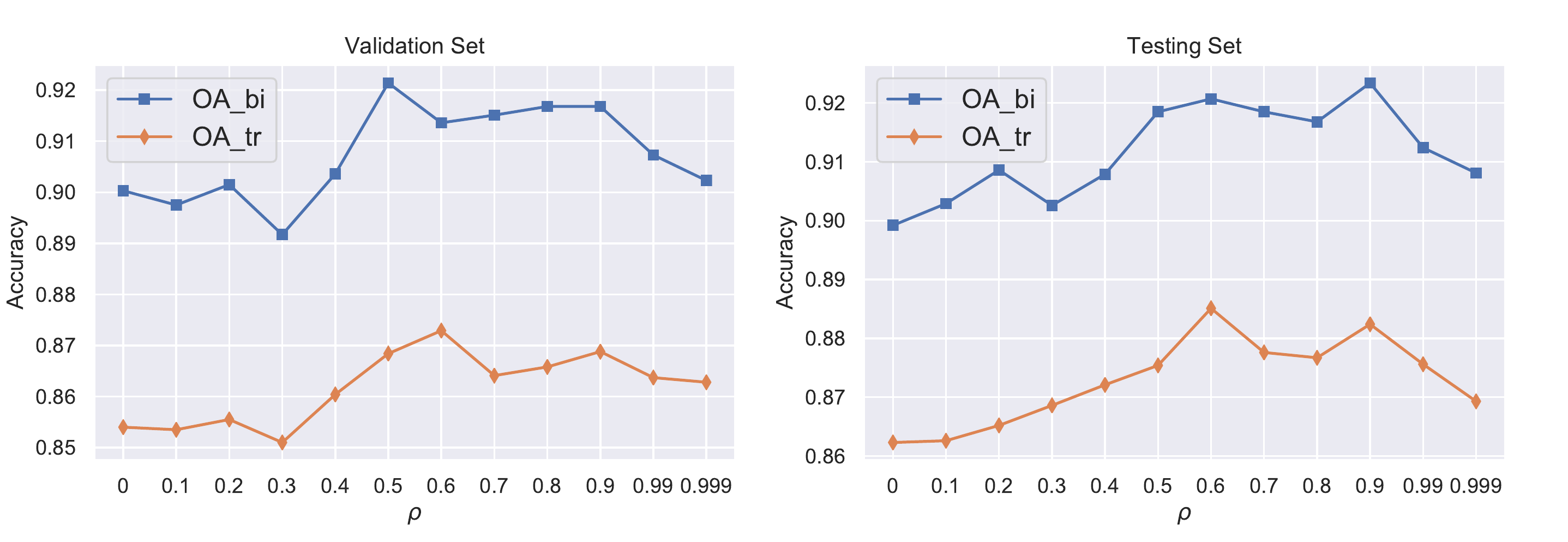}
	\caption{The influence of the momentum parameter $\rho$ in Eq.(\ref{xy_cov})}
	\label{fig_rho}
\end{figure}
\par In further, we evaluated the influence of the momentum parameter $\rho$ on Wuhan dataset in Fig~\ref{fig_rho}. We firstly noted that the performance of $\rho=0$ is worse than others, which demonstrates the necessity of the adaptive estimation for the covariance. As $\rho\rightarrow 0.6$, the scene change detection accuracies increased. But when $\rho$ increased in $[0.9, 0.99, 0.999]$, the accuracies began to decrease, which is accounted in \cite{Wang2016b} that the estimated covariance is not adapted to DNN's outputs as $\rho\rightarrow 1$.
\section{Conclusion}
\label{Conclusion}
\par In this work, we proposed CorrFusionNet to perform multi-temporal scene classification and scene change detection for bi-temporal imagery. CorrFusionNet starts with extracting deep latent feature representations of bi-temporal input imagery. Then the extracted features will be projected into a lower dimensional feature space. A proposed CorrFusion module is employed to compute the temporal correlation and perform the cross-temporal fusion based on the projected features and computed correlation. The scene classification and change detection results will be obtained with softmax layers. The experimental results on a new large-scale scene dataset demonstrated CorrFusionNet could overpass other scene change detection methods.
\par In view of the independence of the two branch convolutional modules, the proposed CorrFusion module could also be easily adapted to cope with multi-source/multi-view classification problems. Besides, except multi-temporal scene classification/ scene change detection problem, the design of CorrFusion module could also be generalized to other multi-temporal problems as a method of enhancing the multi-temporal feature representation.

\section*{Acknowledgment}
The authors would like to thank...
\ifCLASSOPTIONcaptionsoff
	\newpage
\fi


\bibliographystyle{IEEEtran}
\bibliography{IEEEabrv,corrfusion.bbl}

\end{document}